# Probabilistic Deduction with
# Conditional Constraints over Basic Events

**Thomas Lukasiewicz**                                     LUKASIEWICZ@INFORMATIK.UNI-GIESSEN.DE
*Institut für Informatik, Universität Gießen*
*Arndtstraße 2, D-35392 Gießen, Germany*

## Abstract

We study the problem of probabilistic deduction with conditional constraints over basic events. We show that globally complete probabilistic deduction with conditional constraints over basic events is NP-hard. We then concentrate on the special case of probabilistic deduction in conditional constraint trees. We elaborate very efficient techniques for globally complete probabilistic deduction. In detail, for conditional constraint trees with point probabilities, we present a local approach to globally complete probabilistic deduction, which runs in linear time in the size of the conditional constraint trees. For conditional constraint trees with interval probabilities, we show that globally complete probabilistic deduction can be done in a global approach by solving nonlinear programs. We show how these nonlinear programs can be transformed into equivalent linear programs, which are solvable in polynomial time in the size of the conditional constraint trees.

## 1. Introduction

Dealing with uncertain knowledge plays an important role in knowledge representation and reasoning. There are many different formalisms and methodologies for handling uncertainty. Most of them are directly or indirectly based on probability theory.

In this paper, we focus on probabilistic deduction with conditional constraints over basic events (that is, interval restrictions for conditional probabilities of elementary events). The considered probabilistic deduction problems consist of a probabilistic knowledge base and a probabilistic query. We give a classical example. As a probabilistic knowledge base, we may take the probabilistic knowledge that all ostriches are birds, that the probability of Tweety being a bird is greater than 0.90, and that the probability of Tweety being an ostrich provided she is a bird is greater than 0.80. As a probabilistic query, we may now wonder about the entailed greatest lower and least upper bound for the probability that Tweety is an ostrich. The solution to this probabilistic deduction problem is 0.72 for the entailed greatest lower bound and 1.00 for the entailed least upper bound.

More generally, probabilistic deduction with conditional constraints over propositional events can be done in a global approach by linear programming or in a local approach by the iterative application of inference rules. Note that it is immediately NP-hard, since it generalizes the satisfiability problem for classical propositional logic (see Section 2.2).

Research on the global approach spread in particular after the important work on probabilistic logic by Nilsson (1986) (see also the work by Paaß, 1988). The main focus was on analyzing the computational complexity of satisfiability and entailment in probabilistic logic and on developing efficient linear programming algorithms for these problems.

Georgakopoulos et al. (1988) show that the satisfiability problem in probabilistic logic is NP-complete and propose to apply column generation techniques for its processing. This approach was further developed by Kavvadias and Papadimitriou (1990), Jaumard et al. (1991), Andersen and Hooker (1994), and Hansen et al. (1995). In particular, Jaumard et al. (1991) report promising experimental results on the efficiency in special cases of probabilistic satisfiability and entailment. Moreover, Kavvadias and Papadimitriou (1990) and Jaumard et al. (1991) identify special cases of probabilistic satisfiability that can be solved in polynomial time. Other work on the global approach concentrates on reducing the number of linear constraints (Luo et al. 1996) and the number of variables (Lukasiewicz, 1997). Finally, Fagin et al. (1992) present a sound and complete axiom system for reasoning about probabilities that are expressed by linear inequalities over propositional events. They show that the satisfiability problem in this quite expressive framework is still NP-complete.

In early work, Dubois and Prade (1988) use inference rules to model default reasoning with imprecise numerical and fuzzy quantifiers. For this reason, subsequent research on inference rules especially aims at analyzing patterns of human commonsense reasoning (Dubois et al. 1990, 1993; Amarger et al. 1991; Thöne, 1994; Thöne et al. 1995). Frisch and Haddawy (1994) discuss the use of inference rules for deduction in probabilistic logic. Recent work on inference rules concentrates on integrating probabilistic knowledge into description logics (Heinsohn, 1994) and on analyzing the interplay between taxonomic and probabilistic deduction (Lukasiewicz 1998a, 1999a).

We now summarize the main characteristics of the global and the local approach.

The global approach can be performed within quite rich probabilistic languages (Fagin et al., 1992). Crucially, probabilistic deduction by linear programming is globally complete (that is, it really provides the requested tightest bounds entailed by the whole probabilistic knowledge base). However, a main drawback of the global approach is that it generally does not provide useful explanatory information on the deduction process. Finally, results on the special-case tractability of global approaches are driven by the technical possibilities of linear programming techniques and not by the needs of artificial intelligence applications. Hence, they do not seem to be very useful in the artificial intelligence context.

A main advantage of the local approach is that the deduced results can be explained in a natural way by the sequence of applied inference rules (Amarger et al. 1991; Frisch & Haddawy, 1994). However, the iterative application of inference rules is generally restricted to quite narrow probabilistic languages. Moreover, it is very rarely and only within very restricted languages globally complete (Frisch and Haddawy, 1994, give an example of globally complete local probabilistic deduction in a very restricted framework). Finally, as far as the computational complexity is concerned, there are very few experimental and theoretical results on the special-case tractability of local approaches.

The main motivating idea of this paper is to elaborate efficient local techniques for globally complete probabilistic deduction. Inspired by previous work on inference rules, we focus our research on the language of conditional constraints over basic events:

Dubois and Prade (1988) study the chaining of two bidirectional conditional constraints over basic events ("quantified syllogism rule") and some of its special cases. Dubois et al. (1990) additionally discuss probabilistic deductions about conjunctions of basic events. Furthermore, they describe the open problem of probabilistic deduction along a chain of more than two bidirectional conditional constraints over basic events. In later work, Dubois et al. (1993) use a qualitative version of the "quantified syllogism rule" in an approach to reasoning with linguistic quantifiers. Amarger et al. (1991) propose to apply the "quantified syllogism rule" and the "generalized Bayes' rule" to sets of bidirectional conditional constraints over basic events. They report promising experimental results on the global completeness and the computational complexity of the presented deduction technique. However, this deduction technique is generally not globally complete. Thöne (1994) examines trees of bidirectional conditional constraints over basic events. He presents a linear-time deduction technique that is based on a system of inference rules and that computes certain logically entailed greatest lower bounds (in the technical notions of this paper, which will be defined below, tight lower answers to conclusion-restricted queries are computed).

As a first contribution of this paper, we show that globally complete probabilistic deduction with conditional constraints over basic events is NP-hard. It is surprising that this quite restricted class of probabilistic deduction problems is still computationally so difficult. Hence, it is unlikely that there is an algorithm that efficiently solves all probabilistic deduction problems with conditional constraints over basic events. However, we can still hope that there are efficient special-case, average-case, or approximation algorithms.

In this paper, we then elaborate efficient special-case algorithms. In detail, we concentrate on probabilistic deduction in conditional constraint trees. It is an interesting subclass of all probabilistic deduction problems with conditional constraints over basic events. Conditional constraint trees are undirected trees with basic events as nodes and with bidirectional conditional constraints over basic events as edges between the nodes (that is, deduction in conditional constraint trees is a generalization of deduction along a chain of bidirectional conditional constraints over basic events). Like Bayesian networks, conditional constraint trees represent a well-structured probabilistic knowledge base. Differently from Bayesian networks, they do not encode any probabilistic independencies.

As a main contribution of this paper, we have the following results. For conditional constraint trees with point probabilities, we present functions for deducing greatest lower and least upper bounds in linear time in the size of the conditional constraint trees. Moreover, for conditional constraint trees with interval probabilities, we show that greatest lower bounds can be deduced in the same way, in linear time in the size of the conditional constraint trees. However, computing least upper bounds turns out to be computationally more difficult. It can be done by solving special nonlinear programs. We show how these nonlinear programs can be transformed into equivalent linear programs. The resulting linear programs have a number of variables and inequalities linear and polynomial, respectively, in the size of the conditional constraint trees. Thus, our way of deducing least upper bounds still runs in polynomial time in the size of the conditional constraint trees, since linear programming runs in polynomial time in the size of the linear programs.

Another important contribution of this paper is related to the question whether to perform probabilistic deduction with conditional constraints by the iterative application of inference rules or by linear programming. On the one hand, the idea of inference rules carries us to very efficient techniques for globally complete probabilistic deduction in conditional constraint trees. In particular, the considered deduction problems generalize patterns of commonsense reasoning. However, on the other hand, the corresponding proofs of soundness and global completeness are technically quite complex. Hence, it seems unlikely that the results of this work can be extended to significantly more general probabilistic deduction problems. Note that a companion paper (1998a, 1999a) reports similar limits of the local approach in probabilistic deduction under taxonomic knowledge.

The rest of this paper is organized as follows. In Section 2, we formulate the probabilistic deduction problems considered in this work. Section 3 focuses on the probabilistic satisfiability of conditional constraint trees. Section 4 deals with globally complete probabilistic deduction in exact and general conditional constraint trees. In Section 5, we give a comparison with Bayesian networks. Section 6 summarizes the main results of this work.

## 2. Formulating the Probabilistic Deduction Problem

In this section, we introduce the syntactic and semantic notions related to probabilistic knowledge in general and to conditional constraint trees in particular.

### 2.1 Probabilistic Knowledge

Before focusing on the details of conditional constraint trees, we give a general introduction to the kind of probabilistic knowledge considered in this work. We deal with conditional constraints over propositional events. They represent interval restrictions for conditional probabilities of propositional events. Note that the formal background introduced in this section is commonly accepted in the literature (see especially the work by Frisch and Haddawy, 1994, for other work in the same spirit).

We assume a nonempty and finite set of *basic events* $\mathcal{B} = \{B_1, B_2, \ldots, B_n\}$. The set of *conjunctive events* $\mathcal{C}_\mathcal{B}$ is the closure of $\mathcal{B}$ under the Boolean operation $\wedge$. We abbreviate the conjunctive event $C \wedge D$ by $CD$. The set of *propositional events* $\mathcal{G}_\mathcal{B}$ is the closure of $\mathcal{B}$ under the Boolean operations $\wedge$ and $\neg$. We abbreviate the propositional events $G \wedge H$ and $\neg G$ by $GH$ and $\overline{G}$, respectively. The *false event* $B_1 \wedge \neg B_1$ and the *true event* $\neg(B_1 \wedge \neg B_1)$ are abbreviated by $\perp$ and $\top$, respectively. *Conditional constraints* are expressions of the form $(H|G)[u_1, u_2]$ with real numbers $u_1, u_2 \in [0, 1]$ and propositional events $G$ and $H$. In the conditional constraint $(H|G)[u_1, u_2]$, we call $G$ the *premise* and $H$ the *conclusion*.

To define probabilistic interpretations of propositional events and of conditional constraints, we introduce atomic events and the binary relation $\Rightarrow$ between atomic and propositional events. The set of *atomic events* $\mathcal{A}_\mathcal{B}$ is defined by $\mathcal{A}_\mathcal{B} = \{E_1 E_2 \cdots E_n \mid E_i = B_i \text{ or } E_i = \overline{B}_i \text{ for all } i \in [1:n]\}$. Note that each atomic event can be interpreted as a possible world (which corresponds to a mapping from $\mathcal{B}$ to $\{\mathbf{true}, \mathbf{false}\}$). For all atomic events $A$ and all propositional events $G$, let $A \Rightarrow G$ iff $A\overline{G}$ is a propositional contradiction.

A *probabilistic interpretation* $Pr$ is a mapping from $\mathcal{A}_\mathcal{B}$ to $[0, 1]$ such that all $Pr(A)$ with $A \in \mathcal{A}_\mathcal{B}$ sum up to 1. $Pr$ is extended in a well-defined way to propositional events $G$ by: $Pr(G)$ is the sum of all $Pr(A)$ with $A \in \mathcal{A}_\mathcal{B}$ and $A \Rightarrow G$. $Pr$ is extended to conditional constraints by: $Pr \models (H|G)[u_1, u_2]$ iff $u_1 \cdot Pr(G) \leq Pr(GH) \leq u_2 \cdot Pr(G)$.

Note that conditional constraints characterize conditional probabilities of events, rather than *probabilities of conditional events* (Coletti, 1994; Gilio & Scozzafava, 1994). Note also that $Pr(G) = 0$ always entails $Pr \models (H|G)[u_1, u_2]$. This semantics of conditional probability statements is also assumed by Halpern (1990) and by Frisch and Haddawy (1994).

The notions of models, satisfiability, and logical consequence for conditional constraints are defined in the classical way. A probabilistic interpretation $Pr$ is a *model* of a conditional constraint $(H|G)[u_1, u_2]$ iff $Pr \models (H|G)[u_1, u_2]$. $Pr$ is a *model* of a set of conditional constraints $KB$, denoted $Pr \models KB$, iff $Pr$ is a model of all $(H|G)[u_1, u_2] \in KB$. $KB$ is *satisfiable* iff a model of $KB$ exists. $(H|G)[u_1, u_2]$ is a *logical consequence* of $KB$, denoted $KB \models (H|G)[u_1, u_2]$, iff each model of $KB$ is also a model of $(H|G)[u_1, u_2]$.

For a conditional constraint $(H|G)[u_1, u_2]$ and a set of conditional constraints $KB$, let $\boldsymbol{u}$ denote the set of all real numbers $u \in [0, 1]$ for which there exists a model $Pr$ of $KB$ with $u \cdot Pr(G) = Pr(GH)$ and $Pr(G) > 0$. Now, we easily verify that $(H|G)[u_1, u_2]$ is a logical consequence of $KB$ iff $u_1 \leq \inf \boldsymbol{u}$ and $u_2 \geq \sup \boldsymbol{u}$.

This observation yields a canonical notion of tightness for logical consequences of conditional constraints. The conditional constraint $(H|G)[u_1, u_2]$ is a *tight logical consequence* of $KB$, denoted $KB \models_{tight} (H|G)[u_1, u_2]$, iff $u_1 = \inf \boldsymbol{u}$ and $u_2 = \sup \boldsymbol{u}$.

The set $\boldsymbol{u}$ is a closed interval in the real line (Frisch & Haddawy, 1994). Note that for $\boldsymbol{u} = \emptyset$, we canonically define $\inf \boldsymbol{u} = \max [0, 1] = 1$ and $\sup \boldsymbol{u} = \min [0, 1] = 0$. Thus, $\boldsymbol{u} = \emptyset$ iff $KB \models (G|\top)[0, 0]$ iff $KB \models_{tight} (H|G)[1, 0]$ iff $KB \models (H|G)[u_1, u_2]$ for all $u_1, u_2 \in [0, 1]$.

Based on the just introduced notion of tight logical consequence, probabilistic deduction problems and their solutions are more formally specified as follows.

A *probabilistic knowledge base* $(\mathcal{B}, KB)$ consists of a set of basic events $\mathcal{B}$ and a set of conditional constraints $KB$ over $\mathcal{G}_\mathcal{B}$ with $u_1 \leq u_2$ for all $(H|G)[u_1, u_2] \in KB$. A *probabilistic query* to a probabilistic knowledge base $(\mathcal{B}, KB)$ is an expression of the form $\exists (F|E)[x_1, x_2]$ with $E, F \in \mathcal{G}_\mathcal{B}$ and two different variables $x_1$ and $x_2$. Its *tight answer* is the substitution $\sigma = \{x_1/u_1, x_2/u_2\}$ with $u_1, u_2 \in [0, 1]$ such that $KB \models_{tight} (F|E)[u_1, u_2]$ (we call $\sigma_1 = \{x_1/u_1\}$ the *tight lower answer* and $\sigma_2 = \{x_2/u_2\}$ the *tight upper answer*). A *correct answer* is a substitution $\sigma = \{x_1/u_1, x_2/u_2\}$ with $u_1, u_2 \in [0, 1]$ such that $KB \models (F|E)[u_1, u_2]$.

Finally, we define the notions of soundness and of completeness related to inference rules and to techniques for probabilistic deduction. An inference rule $KB \vdash (H|G)[u_1, u_2]$ is *sound* iff $KB \models (H|G)[u_1, u_2]$, where $(H|G)[u_1, u_2]$ is a conditional constraint and $KB$ is a set of conditional constraints. It is sound and *locally complete* iff $KB \models_{tight} (H|G)[u_1, u_2]$. A technique for probabilistic deduction is *sound* iff for a set of probabilistic queries $\mathcal{Q}$ iff it computes a correct answer to any given query from $\mathcal{Q}$. It is sound and *globally complete* for $\mathcal{Q}$ iff it computes the tight answer to any given query from $\mathcal{Q}$.

## 2.2 Computational Complexity

In the framework of conditional constraints over propositional events, the optimization problem of computing the tight answer to a probabilistic query is immediately NP-hard, since it generalizes the satisfiability problem for classical propositional logic (the NP-complete problem of deciding whether a propositional formula in conjunctive normal form is satisfiable; see especially the survey by Garey and Johnson, 1979).

Surprisingly, the optimization problem of computing the tight answer to a probabilistic query remains NP-hard even if we just consider conditional constraints over basic events:

**Theorem 2.1** *The optimization problem of computing the tight answer to a probabilistic query over basic events that is directed to a probabilistic knowledge base over basic events is NP-hard.*

**Proof.** The NP-complete decision problem of graph 3-colorability (Garey & Johnson, 1979) can be polynomially-reduced to the optimization problem of computing the tight answer to a probabilistic query over basic events that is directed to a probabilistic knowledge base over basic events. The proof follows similar lines to the proof of NP-hardness of 2PSAT given by Georgakopoulos et al. (1988).

Let $(V, E)$ be a finite undirected graph. We construct a probabilistic knowledge base $(\mathcal{B}, KB)$ as follows. We initialize $(\mathcal{B}, KB)$ with $(\{B\}, \emptyset)$. For each node $v \in V$, we increase $\mathcal{B}$ by the new basic events $B_v^1$, $B_v^2$, and $B_v^3$. For each node $v \in V$ and for each $i \in \{1, 2, 3\}$, we increase $KB$ by $(B|B_v^i)[1, 1]$ and $(B_v^i|B)[1/3, 1/3]$. For each node $v \in V$ and for each $i, j \in \{1, 2, 3\}$ with $i < j$, we increase $KB$ by $(B_v^j|B_v^i)[0, 0]$. For each edge $\{u, v\} \in E$ and for each $i \in \{1, 2, 3\}$, we increase $KB$ by $(B_v^i|B_u^i)[0, 0]$. It is easy to see that the probabilistic knowledge base $(\mathcal{B}, KB)$ can be constructed in polynomial time in the size of $(V, E)$.

Now, we show that $(V, E)$ is 3-colorable iff $\{x_1/1, \, x_2/1\}$ is the tight answer to the probabilistic query $\exists(B|B)[x_1, x_2]$ to $(\mathcal{B}, KB)$, or equivalently, iff $KB$ is satisfiable:

If $(V, E)$ is 3-colorable, then there exists a mapping $c_1$ from $V$ to $\{1, 2, 3\}$ with $c_1(u) \neq c_1(v)$ for all edges $\{u, v\} \in E$. Thus, if $\pi$ is a cyclic permutation of the members in $\{1, 2, 3\}$ and if $c_2, c_3 \colon V \to \{1, 2, 3\}$ are defined by $c_2(v) = \pi(c_1(v))$ and $c_3(v) = \pi(c_2(v))$ for all nodes $v \in V$, then also $c_2(u) \neq c_2(v)$ and $c_3(u) \neq c_3(v)$ for all edges $\{u, v\} \in E$. For $j \in \{1, 2, 3\}$, let $A_j \in \mathcal{A}_{\mathcal{B}}$ such that $A_j \Rightarrow B$ and $A_j \Rightarrow B_v^i$ iff $c_j(v) = i$ for all nodes $v \in V$ and $i \in \{1, 2, 3\}$. If $Pr \colon \mathcal{A}_{\mathcal{B}} \to [0, 1]$ is defined by $Pr(A) = 1/3$ for all $A \in \{A_1, A_2, A_3\}$ and by $Pr(A) = 0$ for all $A \in \mathcal{A}_{\mathcal{B}} \setminus \{A_1, A_2, A_3\}$, then $Pr$ is a model of $KB$.

Conversely, if there is a model $Pr$ of $KB$, then there is an atomic event $A \in \mathcal{A}_{\mathcal{B}}$ with $Pr(A) > 0$. Thus, if $c \colon V \to \{1, 2, 3\}$ is defined by $c(v) = i$ iff $A \Rightarrow B_v^i$ for all nodes $v \in V$, then $c(u) \neq c(v)$ for all edges $\{u, v\} \in E$. Hence, $(V, E)$ is 3-colorable. □

Hence, it is unlikely that there is an efficient algorithm for computing the tight answer to all probabilistic queries over basic events that are directed to any given probabilistic knowledge base over basic events. However, there may still be efficient algorithms for solving more specialized probabilistic deduction problems.

The rest of this work deals with probabilistic deduction in conditional constraint trees. The next section provides a motivating example, which gives evidence of the practical importance of this kind of probabilistic deduction problems.

## 2.3 Motivating Example

A senior student in mathematics describes her experience about being successful at the university as follows. The success of a student (su) is influenced by how well-informed (wi) and how well-prepared (wp) the student is. Well-informedness can be reached by interviewing professors (pr) or by asking senior students (st). Being well-prepared is influenced by how much time is invested in books (bo), exercises (ex), and hobbies (ho).

It is estimated that the probability of a student being successful given she is well-informed lies between 0.60 and 0.70, that the probability of a student being well-informed given she is successful is greater than 0.85, that the probability of a student being successful given she is well-prepared is greater than 0.95, and that the probability of a student being well-prepared given she is successful is greater than 0.95.

This probabilistic knowledge completed by further probabilistic estimations is given by the probabilistic knowledge base $(\mathcal{B}, KB)$ in Fig. 1, where $\mathcal{B}$ is the set of nodes $\{$su, wi, wp, pr, st, bo, ex, ho} and $KB$ is the least set of conditional constraints that contains $(Y|X)[u_1, u_2]$ for each arrow from $X$ to $Y$ labeled with $u_1, u_2$.

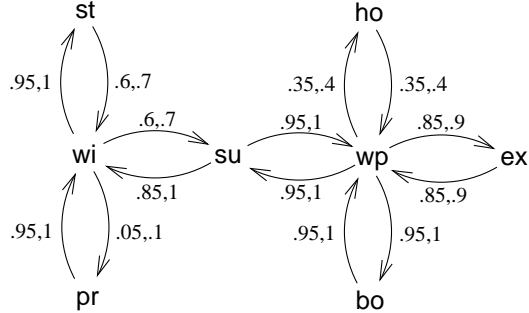

Figure 1: A Conditional Constraint Tree

We may wonder whether it is useful for being successful at the university to interview the professors, to study on books, to spend the time on one's hobbies, or to do both studying on books and spending the time on one's hobbies. This can be expressed by the probabilistic queries $\exists(\mathsf{su}|\mathsf{pr})[x_1, x_2]$, $\exists(\mathsf{su}|\mathsf{bo})[x_1, x_2]$, $\exists(\mathsf{su}|\mathsf{ho})[x_1, x_2]$, and $\exists(\mathsf{su}|\mathsf{bo}\,\mathsf{ho})[x_1, x_2]$, which yield the tight answers $\{x_1/0.00,\ x_2/1.00\}$, $\{x_1/0.90, x_2/1.00\}$, $\{x_1/0.30, x_2/0.46\}$, and $\{x_1/0.71, x_2/1.00\}$, respectively.

We may wonder whether successful students at the university interviewed their professors, whether they studied on books, whether they spent their time with their hobbies, or whether they both studied on books and spent their time with their hobbies. This can be expressed by the probabilistic queries $\exists(\mathsf{pr}|\mathsf{su})[x_1, x_2]$, $\exists(\mathsf{bo}|\mathsf{su})[x_1, x_2]$, $\exists(\mathsf{ho}|\mathsf{su})[x_1, x_2]$, and $\exists(\mathsf{bo}\,\mathsf{ho}|\mathsf{su})[x_1, x_2]$, which yield the tight answers $\{x_1/0.00,\ x_2/0.17\}$, $\{x_1/0.90, x_2/1.00\}$, $\{x_1/0.30, x_2/0.45\}$, and $\{x_1/0.25, x_2/0.45\}$, respectively.

### 2.4 Conditional Constraint Trees

We formally define conditional constraint trees and queries to conditional constraint trees. We provide some additional examples, which are subsequently used as running examples.

A *(general) conditional constraint tree* is a probabilistic knowledge base $(\mathcal{B}, KB)$ for which an undirected tree (a singly connected undirected graph) $(\mathcal{B}, \leftrightarrow)$ exists such that $KB$ contains exactly one pair of conditional constraints $(B|A)[u_1, u_2]$ and $(A|B)[v_1, v_2]$ with $u_1, v_1 > 0$ for each pair of adjacent nodes $A$ and $B$ (note that $\mathcal{B} = \{B\}$ implies $KB = \emptyset$). A basic event $B \in \mathcal{B}$ is called a *leaf* in $(\mathcal{B}, KB)$ iff it has exactly one neighbor in $(\mathcal{B}, \leftrightarrow)$. A conditional constraint tree is *exact* iff $u_1 = u_2$ for all $(B|A)[u_1, u_2] \in KB$.

A *query to a conditional constraint tree* is a probabilistic query $\exists(F|E)[x_1, x_2]$ with two conjunctive events $E$ and $F$ that are disjoint in their basic events and such that all paths from a basic event in $E$ to a basic event in $F$ have at least one basic event in common. A query $\exists(F|E)[x_1, x_2]$ to a conditional constraint tree is *premise-restricted* iff $E$ is a basic event. It is *conclusion-restricted* iff $F$ is a basic event. It is *strongly conclusion-restricted* iff $F$ is the only basic event that is contained in all paths from a basic event in $E$ to $F$. It is *complete* iff $EF$ contains exactly the leaves of $(\mathcal{B}, \leftrightarrow)$.

Fig. 2 shows two conditional constraint trees of which the one on the left side is exact. $\exists(\mathsf{STU}|\mathsf{MNQR})[x_1, x_2]$ is a query, while $\exists(\mathsf{MS}|\mathsf{QU})[x_1, x_2]$ is not a query to the conditional constraint trees of Fig. 2. Furthermore, $\exists(\mathsf{STU}|\mathsf{M})[x_1, x_2]$ is a premise-restricted query, $\exists(\mathsf{O}|\mathsf{QRSTU})[x_1, x_2]$ a strongly conclusion-restricted query, and $\exists(\mathsf{QRSTU}|\mathsf{M})[x_1, x_2]$ a premise-restricted complete query to the conditional constraint trees of Fig. 2.

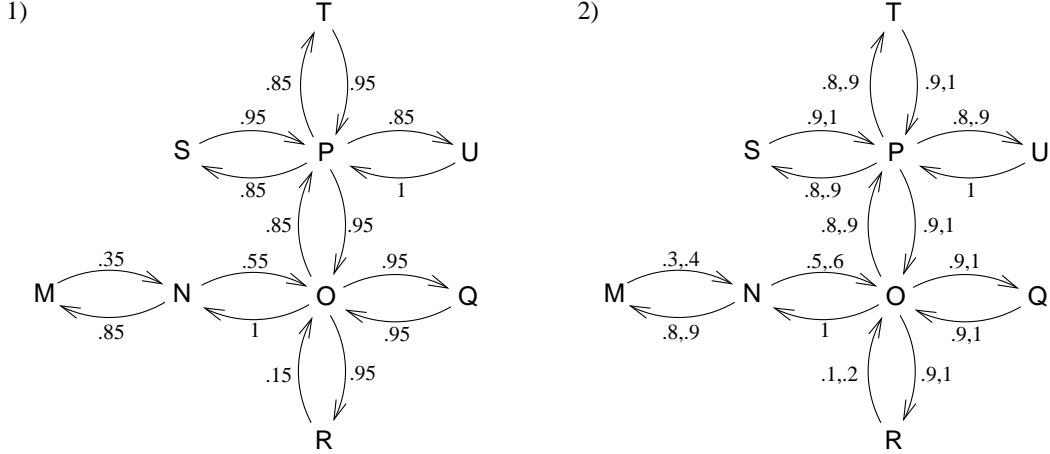

Figure 2: Two Conditional Constraint Trees

For conditional constraint trees $(\mathcal{B}, KB)$, conjunctive events $C$, and basic events $B$, we write $C \Rightarrow B$ iff there exists a path $G_1, G_2, \ldots, G_k$ from a basic event $G_1$ in $C$ to the basic event $G_k = B$ such that $(G_{i+1}|G_i)[1, 1] \in KB$ for all $i \in [1 : k-1]$. We write $B \Rightarrow C$ iff for all paths $G_1, G_2, \ldots, G_k$ from the basic event $G_1 = B$ to a basic event $G_k$ in $C$, it holds $(G_{i+1}|G_i)[1, 1] \in KB$ for all $i \in [1 : k-1]$. That is, the conditions $C \Rightarrow B$ and $B \Rightarrow C$ immediately entail $KB \models (B|C)[1, 1]$ and $KB \models (C|B)[1, 1]$, respectively.

Note that the restriction $u_1, v_1 > 0$ for all $(B|A)[u_1, u_2]$, $(A|B)[v_1, v_2] \in KB$ is just made for technical convenience. The deduction technique of Section 4 can easily be generalized to conditional constraint trees $(\mathcal{B}, KB)$ that satisfy only the restriction $u_1 > 0$ iff $v_1 > 0$ for all $(B|A)[u_1, u_2]$, $(A|B)[v_1, v_2] \in KB$ (Lukasiewicz, 1996).

The restriction that for each query $\exists(F|E)[x_1, x_2]$, all paths from a basic event in $E$ to a basic event in $F$ have at least one basic event in common is crucial for the deduction technique of Section 4. It assures that the problem of computing the tight answer to a complete query can be reduced to the problems of computing the tight answer to a premise-restricted complete query and the tight answer to a strongly conclusion-restricted complete query. Note that this restriction is trivially satisfied by all premise- and conclusion-restricted queries (for example, by all the queries in Section 2.3).

Especially tight answers to conclusion-restricted queries seem to be quite important in practice. They may be used to characterize the probability of uncertain basic events given a collection of basic events that are known with certainty.

## 3. Probabilistic Satisfiability

In this section, we show that conditional constraint trees have the nice property that they are always satisfiable. That is, within conditional constraint trees, the user is prevented from specifying inconsistent probabilistic knowledge.

First, note that conditional constraint trees always have a trivial model in which the probability of the conjunction of all negated basic events is one and in which the probability of all the other atomic events is zero.

The next lemma shows that, given a model $Pr$ of a conditional constraint tree and a real number $s$ from $[0, 1]$, we can construct a new model $Pr_s$ by setting $Pr_s(A) = s \cdot Pr(A)$ for all atomic events $A$ that are different from the conjunction of all negated basic events. Note that $Pr_0$ coincides with the trivial model and that $Pr_1$ is identical to $Pr$. This lemma is crucial for inductively constructing models of conditional constraint trees.

**Lemma 3.1** *Let $(\mathcal{B}, KB)$ be a conditional constraint tree with $\mathcal{B} = \{B_1, B_2, \ldots, B_n\}$. Let $Pr$ be a model of $KB$ and let $s$ be a real number from $[0, 1]$.*
*The mapping $Pr_s \colon \mathcal{A}_\mathcal{B} \to [0, 1]$ with $Pr_s(A) = s \cdot Pr(A)$ for all $A \in \mathcal{A}_\mathcal{B} \setminus \{\overline{B_1}\,\overline{B_2} \cdots \overline{B_n}\}$ and $Pr_s(\overline{B_1}\,\overline{B_2} \cdots \overline{B_n}) = s \cdot Pr(\overline{B_1}\,\overline{B_2} \cdots \overline{B_n}) - s + 1$ is a model of $KB$.*

**Proof.** We easily verify that $Pr_s$ is a probabilistic interpretation. It remains to show that $Pr_s$ is also a model of $KB$. Let $(H|G)[u_1, u_2] \in KB$. Since $Pr$ is a model of $KB$, we have $Pr \models (H|G)[u_1, u_2]$, hence $u_1 \cdot Pr(G) \leq Pr(GH) \leq u_2 \cdot Pr(G)$, and thus also $u_1 s \cdot Pr(G) \leq s \cdot Pr(GH) \leq u_2 s \cdot Pr(G)$. Since neither $\overline{B_1}\,\overline{B_2} \cdots \overline{B_n} \Rightarrow G$ nor $\overline{B_1}\,\overline{B_2} \cdots \overline{B_n} \Rightarrow GH$, we get $u_1 \cdot Pr_s(G) \leq Pr_s(GH) \leq u_2 \cdot Pr_s(G)$ and thus $Pr_s \models (H|G)[u_1, u_2]$. $\square$

Finally, the following theorem shows that conditional constraint trees always have a nontrivial model in which all the basic events have a probability greater than zero.

**Theorem 3.2** *Let $(\mathcal{B}, KB)$ be a conditional constraint tree with $\mathcal{B} = \{B_1, B_2, \ldots, B_n\}$.*
*There is a model $Pr$ of $KB$ with $Pr(B_1 B_2 \cdots B_n) > 0$.*

**Proof.** It is sufficient to show the claim for exact conditional constraint trees. The claim is proved by induction on the number of basic events.

*Basis:* for $(\mathcal{B}, KB) = (\{B\}, \emptyset)$, a model $Pr$ of $KB$ with $Pr(B) > 0$ is given by $\overline{B}, B \mapsto 0, 1$ (note that $\overline{B}, B \mapsto 0, 1$ is an abbreviation for $Pr(\overline{B}) = 0$ and $Pr(B) = 1$).

*Induction:* let $(\mathcal{B}, KB) = (\mathcal{B}_1 \cup \mathcal{B}_2, KB_1 \cup KB_2)$ with two exact conditional constraint trees $(\mathcal{B}_1, KB_1) = (\{B, C\}, \{(C|B)[u, u], (B|C)[v, v]\})$ and $(\mathcal{B}_2, KB_2) = (\{C, D_1, \ldots, D_k\}, KB_2)$ such that $\mathcal{B}_1 \cap \mathcal{B}_2 = \{C\}$. A model $Pr_1$ of $KB_1$ with $Pr_1(BC) > 0$ is given by:

$$\overline{B}\,\overline{C}, B\overline{C}, \overline{B}C, BC \;\mapsto\; \tfrac{uv}{u+v}, \tfrac{v-uv}{u+v}, \tfrac{u-uv}{u+v}, \tfrac{uv}{u+v} \,.$$

By the induction hypothesis, there is a model $Pr_2$ of $KB_2$ (that is defined on the atomic events over $\mathcal{B}_2$) with $Pr_2(CD_1 \cdots D_k) > 0$. By Lemma 3.1, we can assume $Pr_2(C) = Pr_1(C)$. A probabilistic interpretation $Pr$ on the atomic events over $\mathcal{B}$ is now defined by:

$$Pr(A_b A_c A_2) \;=\; \tfrac{Pr_1(A_b A_c) \cdot Pr_2(A_c A_2)}{Pr_2(A_c)}$$

for all atomic events $A_b$, $A_c$, and $A_2$ over $\{B\}$, $\{C\}$, and $\mathcal{B}_2 \setminus \{C\}$, respectively. We easily verify that $Pr(A_b A_c) = Pr_1(A_b A_c)$ and $Pr(A_c A_2) = Pr_2(A_c A_2)$ for all atomic events $A_b$, $A_c$, and $A_2$ over $\{B\}$, $\{C\}$, and $\mathcal{B}_2 \setminus \{C\}$, respectively. Hence, $Pr$ is a model of $KB$. Moreover, $Pr_1(BC) > 0$ and $Pr_2(CD_1 \cdots D_k) > 0$ entails $Pr(BCD_1 \cdots D_k) > 0$. □

## 4. Probabilistic Deduction

In this section, we present techniques for computing tight answers to queries directed to exact and general conditional constraint trees, and we analyze their computational complexity. More precisely, the problem of computing the tight answer to a query is reduced to the problem of computing the tight answer to a complete query. The latter problem is then reduced to the problems of computing the tight answer to a premise-restricted complete query and the tight answer to a strongly conclusion-restricted complete query.

### 4.1 Premise-Restricted Complete Queries

#### 4.1.1 EXACT CONDITIONAL CONSTRAINT TREES

We now focus on the problem of computing tight answers to premise-restricted complete queries that are directed to exact conditional constraint trees.

Let $(\mathcal{B}, KB)$ be an exact conditional constraint tree and let $\exists (F|E)[x_1, x_2]$ be a premise-restricted complete query. To compute the tight answer to $\exists (F|E)[x_1, x_2]$, we start by defining a directed tree (that is, a directed acyclic graph in which each node has exactly one parent, except for the root, which does not have any):

$$A \to B \quad \text{iff} \quad A \leftrightarrow B \text{ and } A \text{ is closer to } E \text{ than } B.$$

This directed tree $(\mathcal{B}, \to)$ is uniquely determined by the conditional constraint tree and the premise-restricted complete query. Fig. 3 shows $(\mathcal{B}, \to)$ for the premise-restricted complete query $\exists (\mathsf{QRSTU}|\mathsf{M})[x_1, x_2]$ to the exact conditional constraint tree in Fig. 2, left side.

Now, the set of nodes $\mathcal{B}$ is partitioned into several strata. The lowest stratum contains only nodes with no children in $(\mathcal{B}, \to)$, the highest stratum contains the nodes with no parents in $(\mathcal{B}, \to)$ (that is, exactly the node of the premise $E$ of the query). Fig. 3 also shows the different strata in our example.

At each node of $(\mathcal{B}, \to)$, we compute certain tightest bounds that are logically entailed by $KB$. More precisely, the tightest bounds at a node $B$ are computed locally, by exploiting the tightest bounds that have previously been computed at the children of $B$. Hence, we iteratively compute the tightest bounds at the nodes of each stratum, starting with the nodes of the lowest stratum and terminating with the nodes of the highest stratum. We distinguish three different ways of computing tightest bounds at a node:

- initialization of a leaf (LEAF),

- chaining of an arrow and a subtree via a common node (CHAINING),

- fusion of subtrees via a common node (FUSION).

Let us consider again the premise-restricted complete query $\exists (\mathsf{QRSTU}|\mathsf{M})[x_1, x_2]$ to the exact conditional constraint tree in Fig. 2, left side. Fig. 4 illustrates the three different ways

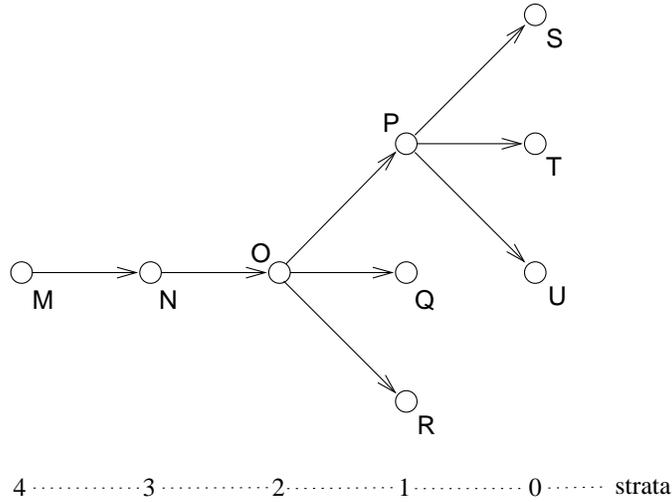

Figure 3: Directed Tree $(\mathcal{B}, \rightarrow)$

of computing tightest bounds at a node (the common nodes for CHAINING and FUSION are filled black). Table 1 shows the greatest lower and the least upper bounds that are computed at each node $B$ of each stratum. More precisely, these bounds are $\alpha_1 = \inf Pr(BD)/Pr(B)$, $\alpha_2 = \sup Pr(BD)/Pr(B)$, $\beta_2 = \sup Pr(\overline{B}D)/Pr(B)$, and $\gamma_2 = \sup Pr(D)/Pr(B)$ subject to $Pr \models KB$ and $Pr(B) > 0$. Table 1 also shows the requested tight answer $\{x_1/0.02,\ x_2/0.17\}$, which is given by the tightest bounds $\alpha_1$ and $\alpha_2$ that are computed at the premise M.

| strata | $B$ | $D$ | $\alpha_1$ | $\alpha_2$ | $\beta_2$ | $\gamma_2$ | |
|---|---|---|---|---|---|---|---|
| | S | S | 1.0000 | 1.0000 | 0.0000 | 1.0000 | (LEAF) |
| 0 | T | T | 1.0000 | 1.0000 | 0.0000 | 1.0000 | (LEAF) |
| | U | U | 1.0000 | 1.0000 | 0.0000 | 1.0000 | (LEAF) |
| | P | S | 0.8500 | 0.8500 | 0.0447 | 0.8947 | (CHAINING) |
| 1 | P | T | 0.8500 | 0.8500 | 0.0447 | 0.8947 | (CHAINING) |
| | P | U | 0.8500 | 0.8500 | 0.0000 | 0.8500 | (CHAINING) |
| | P | STU | 0.5500 | 0.8500 | 0.0000 | 0.8500 | (FUSION) |
| 1 | Q | Q | 1.0000 | 1.0000 | 0.0000 | 1.0000 | (LEAF) |
| | R | R | 1.0000 | 1.0000 | 0.0000 | 1.0000 | (LEAF) |
| | O | STU | 0.4474 | 0.7605 | 0.0447 | 0.7605 | (CHAINING) |
| 2 | O | Q | 0.9500 | 0.9500 | 0.0500 | 1.0000 | (CHAINING) |
| | O | R | 0.9500 | 0.9500 | 5.3833 | 6.3333 | (CHAINING) |
| 2 | O | QRSTU | 0.3474 | 0.7605 | 0.0447 | 0.7605 | (FUSION) |
| 3 | N | QRSTU | 0.1911 | 0.4183 | 0.0246 | 0.4183 | (CHAINING) |
| 4 | M | QRSTU | 0.0169 | 0.1722 | 0.0719 | 0.1722 | (CHAINING) |

Table 1: Locally Computed Tightest Bounds

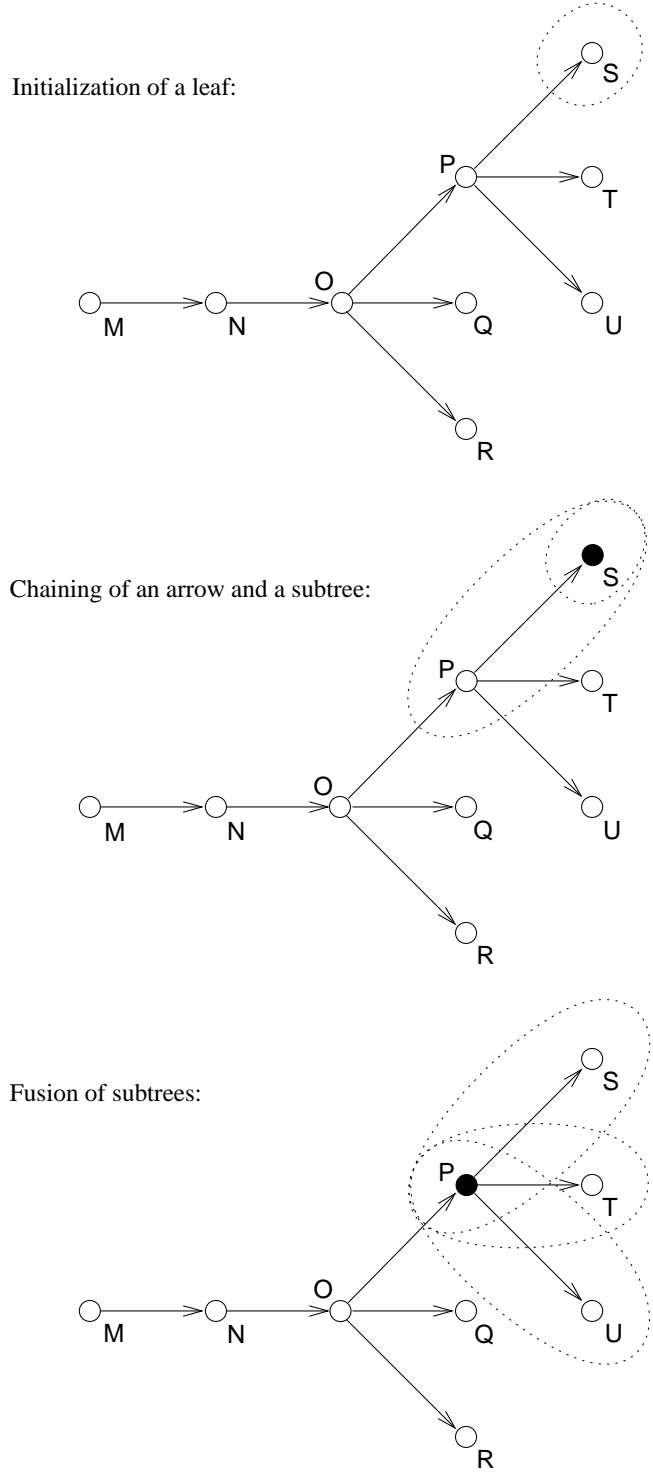

Initialization of a leaf:

Chaining of an arrow and a subtree:

Fusion of subtrees:

Figure 4: Local Computations in $(\mathcal{B}, \rightarrow)$

We now focus on the technical details. We present the functions $H_1^\alpha$, $H_2^\alpha$, $H_2^\beta$, and $H_2^\gamma$, which compute the described greatest lower and least upper bounds. For this purpose, we need the following definitions. Let $Pr(C|B)$ denote $u$ for all $(C|B)[u, u] \in KB$.

A node $B$ is a *leaf* if it does not have any children. For all leaves $B$, let $B^\uparrow = B$. For all the other nodes $B$, let $B^\uparrow$ be the conjunction of all the children of $B$. For all leaves $C$, let $L(C) = C$. For all the other conjunctive events $C$, let $L(C)$ be the conjunction of all the leaves that are in $C$ or that are descendants of a node in $C$.

In the sequel, let $B$ be a node and let $C = B^\uparrow$. The case $C = B$ refers to the initialization of the leaf $B$, the case $C = B_1$ with a node $B_1 \neq B$ to the chaining of the arrow $B \to B_1$ and a subtree via the common node $B_1$, and the case $C = B_1 B_2 \ldots B_k$ with $k > 1$ nodes $B_1, B_2, \ldots, B_k$ to the fusion of $k$ subtrees via the common node $B$.

We define the function $H_1^\alpha$ for computing greatest lower bounds: let $H_1^\alpha(B, C) = \alpha_1$ (note that $\alpha_1$ will coincide with the greatest lower bound of $Pr(BL(C)) / Pr(B)$ subject to $Pr \models KB$ and $Pr(B) > 0$), where $\alpha_1$ in Leaf ($C = B$), Chaining ($C = B_1$), and Fusion ($C = B_1 B_2 \ldots B_k$ with $k > 1$) is given as follows:

Leaf:

$$\alpha_1 \;=\; 1$$

Chaining:

$$\alpha_1 \;=\; \max(0, Pr(C|B) \cdot (1 + \tfrac{H_1^\alpha(C, C^\uparrow) - 1}{Pr(B|C)}))$$

Fusion:

$$\alpha_1 \;=\; \max(0, 1 - k + \sum_{i=1}^{k} H_1^\alpha(B, B_i))$$

To express that $H_1^\alpha$ computes greatest lower bounds, we need the following definitions. Let $\mathcal{B}(B, C)$ comprise $B$, all nodes in $C$ and all descendants of a node in $C$. Let $KB(B, C)$ be the set of all conditional constraints of $KB$ over $\mathcal{B}(B, C)$. Let $Mo(B, C)$ be the set of all models of $KB(B, C)$ that are defined on the atomic events over $\mathcal{B}(B, C)$.

Now, the function $H_1^\alpha$ is sound and globally complete with respect to $B$ and $C$ iff $H_1^\alpha(B, C) = \alpha_1$ is the greatest lower bound of $Pr(BL(C)) / Pr(B)$ subject to $Pr \in Mo(B, C)$ and $Pr(B) > 0$. Thus, the next theorem shows soundness and global completeness of $H_1^\alpha$.

**Theorem 4.1**

*a) For all probabilistic interpretations $Pr \in Mo(B, C)$, it holds $\alpha_1 \cdot Pr(B) \leq Pr(BL(C))$.*

*b) There exists a probabilistic interpretation $Pr \in Mo(B, C)$ with $Pr(B) > 0$, $\alpha_1 \cdot Pr(B) = Pr(BL(C))$, and $Pr(\overline{B}L(C)) = 0$ iff $L(C) \Rightarrow B$.*

**Proof.** The proof is given in full detail in Appendix B. $\square$

Next, we present the functions $H_2^\alpha$, $H_2^\beta$, and $H_2^\gamma$ for computing least upper bounds. Note that $H_2^\alpha$, $H_2^\beta$, and $H_2^\gamma$ show the crucial result that for exact conditional constraint trees, there are local probabilistic deduction techniques that are sound and globally complete.

In detail, let $H_2^\alpha(B,C) = \alpha_2$, $H_2^\beta(B,C) = \beta_2$, and $H_2^\gamma(B,C) = \gamma_2$ (note that $\alpha_2$, $\beta_2$, and $\gamma_2$ will coincide with the least upper bound of $Pr(BL(C))\,/\,Pr(B)$, $Pr(\overline{B}L(C))\,/\,Pr(B)$, and $Pr(L(C))\,/\,Pr(B)$, respectively, subject to $Pr \models KB$ and $Pr(B) > 0$), where $\alpha_2$, $\beta_2$, and $\gamma_2$ in Leaf $(C = B)$, Chaining $(C = B_1)$, and Fusion $(C = B_1 B_2 \ldots B_k$ with $k > 1)$ are given as follows:

Leaf:

$$\alpha_2 = 1$$
$$\beta_2 = 0$$
$$\gamma_2 = 1$$

Chaining:

$$\alpha_2 = \min(1, Pr(C|B) \cdot \tfrac{H_2^\gamma(C,C^\uparrow)}{Pr(B|C)},\ 1 - Pr(C|B) \cdot (1 - \tfrac{H_2^\alpha(C,C^\uparrow)}{Pr(B|C)}),$$
$$Pr(C|B) \cdot (1 + \tfrac{H_2^\beta(C,C^\uparrow)}{Pr(B|C)}))$$
$$\beta_2 = \min(Pr(C|B) \cdot (\tfrac{H_2^\beta(C,C^\uparrow)+1}{Pr(B|C)} - 1),\ Pr(C|B) \cdot \tfrac{H_2^\gamma(C,C^\uparrow)}{Pr(B|C)})$$
$$\gamma_2 = Pr(C|B) \cdot \tfrac{H_2^\gamma(C,C^\uparrow)}{Pr(B|C)}$$

Fusion:

$$\alpha_2 = \min_{i \in [1:k]} H_2^\alpha(B, B_i)$$
$$\beta_2 = \min_{i \in [1:k]} H_2^\beta(B, B_i)$$
$$\gamma_2 = \min(\min_{i \in [1:k]} H_2^\gamma(B, B_i),\ \min_{i,j \in [1:k], i \neq j} (H_2^\alpha(B, B_i) + H_2^\beta(B, B_j)))$$

The functions $H_2^\alpha$, $H_2^\beta$, and $H_2^\gamma$ are sound and globally complete with respect to $B$ and $C$ iff $H_2^\alpha(B,C) = \alpha_2$, $H_2^\beta(B,C) = \beta_2$, and $H_2^\gamma(B,C) = \gamma_2$ are the least upper bounds of $Pr(BL(C))\,/\,Pr(B)$, $Pr(\overline{B}L(C))\,/\,Pr(B)$, and $Pr(L(C))\,/\,Pr(B)$, respectively, subject to $Pr \in Mo(B,C)$ and $Pr(B) > 0$. Hence, the following theorem shows soundness and global completeness of $H_2^\alpha$, $H_2^\beta$, and $H_2^\gamma$ (actually, it shows even more to enable a proof by induction on the recursive definition of $H_2^\alpha$, $H_2^\beta$, and $H_2^\gamma$).

**Theorem 4.2**

a) *For all probabilistic interpretations $Pr \in Mo(B,C)$, it holds $Pr(BL(C)) \leq \alpha_2 \cdot Pr(B)$, $Pr(\overline{B}L(C)) \leq \beta_2 \cdot Pr(B)$, and $Pr(L(C)) \leq \gamma_2 \cdot Pr(B)$.*

b) *There exists a probabilistic interpretation $Pr \in Mo(B,C)$ with $Pr(B) > 0$, $Pr(BL(C)) = \alpha_2 \cdot Pr(B)$, and $Pr(L(C)) = \gamma_2 \cdot Pr(B)$.*

c) *There exists a probabilistic interpretation $Pr \in Mo(B,C)$ with $Pr(B) > 0$, $Pr(\overline{B}L(C)) = \beta_2 \cdot Pr(B)$, and $Pr(L(C)) = \gamma_2 \cdot Pr(B)$.*

**Proof.** The proof is given in full detail in Appendix B. □

Note that Theorem 4.2 also shows that $H_2^\gamma(B, B_i) \leq H_2^\alpha(B, B_i) + H_2^\beta(B, B_i)$ for all $i \in [1{:}k]$. Thus, the expression $\min_{i,j \in [1:k], i \neq j}(H_2^\alpha(B, B_i) + H_2^\beta(B, B_j))$ in the definition of $\gamma_2$ in Fusion can be replaced by $\alpha_2 + \beta_2$ for an increased efficiency in computing $\gamma_2$ by exploiting the already computed values of $\alpha_2$ and $\beta_2$.

Briefly, by Theorems 4.1 and 4.2, the tight answer to the premise-restricted complete query $\exists (F|E)[x_1, x_2]$ is given by $\{x_1 / H_1^\alpha(E, E^\uparrow), \ x_2 / H_2^\alpha(E, E^\uparrow)\}$.

### 4.1.2 Conditional Constraint Trees

We now focus on computing the tight answer to premise-restricted complete queries to general conditional constraint trees. In the sequel, let $(\mathcal{B}, KB)$ be a conditional constraint tree and let $\exists (F|E)[x_1, x_2]$ be a premise-restricted complete query.

We may think that the local deduction technique for exact conditional constraint trees of Section 4.1.1 can easily be generalized to conditional constraint trees. In fact, this is true as far as the computation of greatest lower bounds is concerned. However, the computation of least upper bounds cannot be generalized that easily from exact conditional constraint trees to conditional constraint trees. More precisely, generalizing the computation of least upper bounds results in solving nonlinear programs. These nonlinear programs and our way to solve them are illustrated by the following chaining example.

Let the conditional constraint tree $(\mathcal{B}, KB)$ be given by $\mathcal{B} = \{\mathsf{M}, \mathsf{N}, \mathsf{O}, \mathsf{P}\}$ and $KB = \{(\mathsf{N}|\mathsf{M})[u_1, u_2], (\mathsf{M}|\mathsf{N})[v_1, v_2], (\mathsf{O}|\mathsf{N})[x_1, x_2], (\mathsf{N}|\mathsf{O})[y_1, y_2], (\mathsf{P}|\mathsf{O})[r_1, r_2], (\mathsf{O}|\mathsf{P})[s_1, s_2]\}$. Let us consider the premise-restricted complete query $\exists (\mathsf{P}|\mathsf{M})[z_1, z_2]$.

By Theorem 4.2 and some straightforward arithmetic transformations, the requested least upper bound is the maximum of $z$ subject to $u \in [u_1, u_2]$, $v \in [v_1, v_2]$, $x \in [x_1, x_2]$, $y \in [y_1, y_2]$, $r \in [r_1, r_2]$, $s \in [s_1, s_2]$, and the nonlinear inequalities in (1) to (5):

$$\text{(1)} \qquad z \leq 1$$
$$\text{(2)} \qquad z \leq 1 - u + \frac{u}{v} - \frac{ux}{v} + \frac{uxr}{vy}$$
$$\text{(3)} \qquad z \leq 1 - u + \frac{ux}{v} - \frac{uxr}{vy} + \frac{uxr}{vys}$$
$$\text{(4)} \qquad z \leq u - \frac{ux}{v} + \frac{ux}{vy} - \frac{uxr}{vy} + \frac{uxr}{vys}$$
$$\text{(5)} \qquad z \leq \frac{uxr}{vys}$$

In this system of nonlinear inequalities, all upper bounds of $z$ are monotonically decreasing in $v$, $y$, and $s$. Hence, we can equivalently maximize $z$ subject to $u \in [u_1, u_2]$, $x \in [x_1, x_2]$, $r \in [r_1, r_2]$, and the nonlinear inequalities in (6) to (10):

$$\text{(6)} \qquad z \leq 1$$
$$\text{(7)} \qquad z \leq 1 - u + \frac{u}{v_1} - \frac{ux}{v_1} + \frac{uxr}{v_1 y_1}$$
$$\text{(8)} \qquad z \leq 1 - u + \frac{ux}{v_1} - \frac{uxr}{v_1 y_1} + \frac{uxr}{v_1 y_1 s_1}$$
$$\text{(9)} \qquad z \leq u - \frac{ux}{v_1} + \frac{ux}{v_1 y_1} - \frac{uxr}{v_1 y_1} + \frac{uxr}{v_1 y_1 s_1}$$
$$\text{(10)} \qquad z \leq \frac{uxr}{v_1 y_1 s_1}$$

For example, the requested least upper bound for $u_1 = u_2 = u$ and $x_1 = x_2 = x$ is shown in Fig. 5 for $u, x \in [0, 1]$, $r_1 = r_2 = 0.15$, $v_1 = 0.8$, $y_1 = 0.8$, and $s_1 \in \{0.05, 0.1\}$. The requested least upper bound for $u_1 < u_2$ or $x_1 < x_2$ is the maximum value over $[u_1, u_2] \times [x_1, x_2]$.

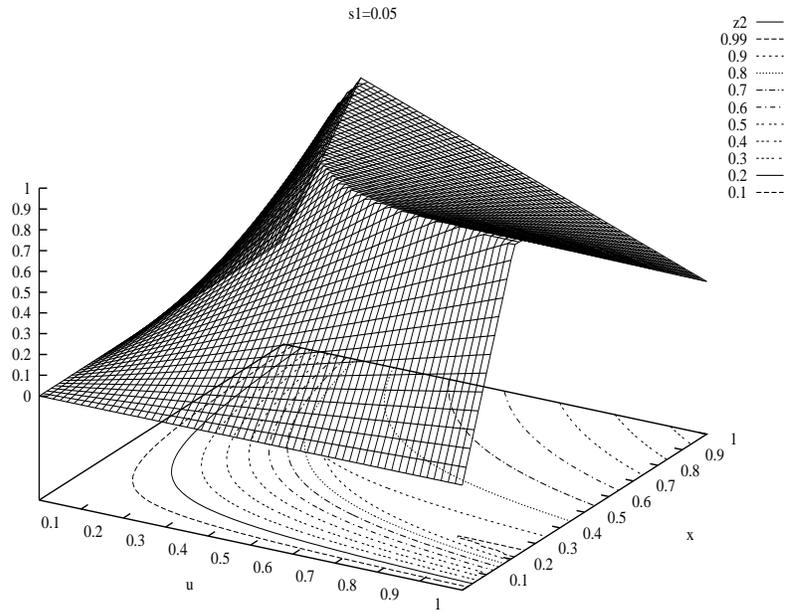

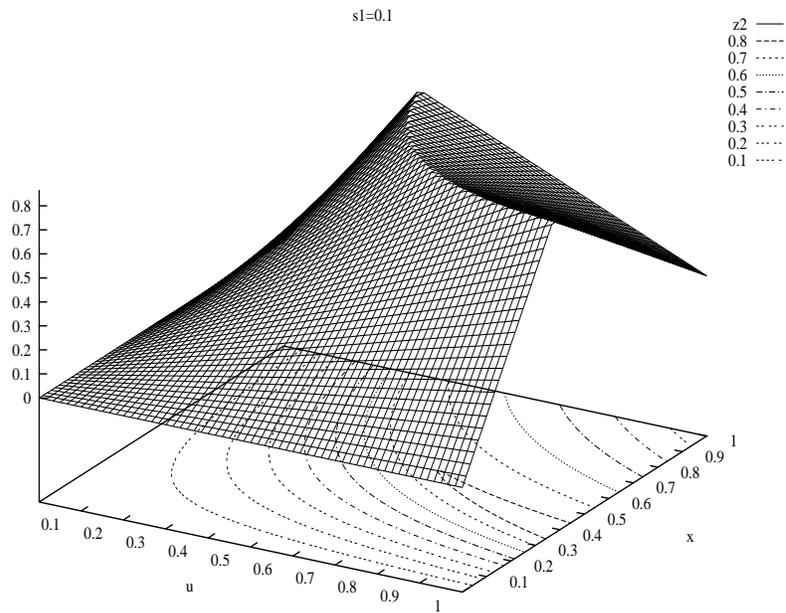

Figure 5: Least Upper Bound $z_2$ in the Chaining Example

We now transform this nonlinear program into an equivalent linear program (by replacing 1, $u$, $ux$, and $uxr$ by the new variables $x_M$, $x_N$, $x_O$, and $x_P$, respectively). More precisely, the maximum of $z$ subject to $u \in [u_1, u_2]$, $x \in [x_1, x_2]$, $r \in [r_1, r_2]$, and the nonlinear inequalities in (6) to (10) coincides with the maximum of $z$ subject to the following system of linear inequalities over $z$ and $x_B \geq 0$ $(B \in \mathcal{B})$:

$$
\begin{aligned}
z &\leq x_M \\
z &\leq x_M + \tfrac{1-v_1}{v_1} \cdot x_N - \tfrac{y_1}{v_1 y_1} \cdot x_O + \tfrac{s_1}{v_1 y_1 s_1} \cdot x_P \\
z &\leq x_M - \tfrac{v_1}{v_1} \cdot x_N + \tfrac{y_1}{v_1 y_1} \cdot x_O + \tfrac{1-s_1}{v_1 y_1 s_1} \cdot x_P \\
z &\leq \quad\quad \tfrac{v_1}{v_1} \cdot x_N + \tfrac{1-y_1}{v_1 y_1} \cdot x_O + \tfrac{1-s_1}{v_1 y_1 s_1} \cdot x_P \\
z &\leq \quad\quad\quad\quad\quad\quad\quad\quad\quad\quad \tfrac{1}{v_1 y_1 s_1} \cdot x_P
\end{aligned}
\qquad
\begin{aligned}
1 &\leq x_M \leq 1 \\
u_1 \cdot x_M &\leq x_N \leq u_2 \cdot x_M \\
x_1 \cdot x_N &\leq x_O \leq x_2 \cdot x_N \\
r_1 \cdot x_O &\leq x_P \leq r_2 \cdot x_O
\end{aligned}
$$

More generally, tight upper answers to premise-restricted complete queries to conditional constraint trees can be computed by solving similar nonlinear programs, which can similarly be transformed into linear programs.

For example, let us consider the premise-restricted complete query $\exists(QRSTU|M)[x_1, x_2]$ to the conditional constraint tree in Fig. 2, right side. The requested least upper bound is the maximum of $x$ subject to the system of linear inequalities in Fig. 6 (we actually generated 72 linear inequalities of which 31 were trivially subsumed by others). Note that the nine variables $x_M$ to $x_U$ correspond to the nine nodes M to U.

$$
\begin{aligned}
x &\leq x_M \\
x &\leq \tfrac{25}{18} x_Q \\
x &\leq \tfrac{25}{2} x_R \\
x &\leq \tfrac{25}{18} x_S \\
x &\leq \tfrac{25}{18} x_T \\
x &\leq \tfrac{25}{18} x_U \\
x &\leq x_N + \tfrac{1}{9} x_P \\
x &\leq x_N + \tfrac{1}{9} x_Q \\
x &\leq x_N + \tfrac{1}{9} x_R \\
x &\leq \tfrac{5}{4} x_O + \tfrac{5}{36} x_P \\
x &\leq \tfrac{5}{4} x_O + \tfrac{5}{36} x_Q \\
x &\leq \tfrac{5}{4} x_O + \tfrac{45}{4} x_R
\end{aligned}
\qquad
\begin{aligned}
x &\leq \tfrac{5}{4} x_P + \tfrac{5}{36} x_Q \\
x &\leq \tfrac{5}{36} x_P + \tfrac{5}{4} x_Q \\
x &\leq \tfrac{5}{36} x_P + \tfrac{5}{4} x_R \\
x &\leq \tfrac{5}{36} x_Q + \tfrac{5}{4} x_R \\
x &\leq x_M - x_N + \tfrac{5}{4} x_O + \tfrac{5}{36} x_R \\
x &\leq x_M + \tfrac{1}{4} x_N - \tfrac{5}{4} x_O + \tfrac{5}{4} x_P \\
x &\leq x_M + \tfrac{1}{4} x_N - \tfrac{5}{4} x_O + \tfrac{5}{4} x_Q \\
x &\leq x_M + \tfrac{1}{4} x_N - \tfrac{5}{4} x_O + \tfrac{5}{4} x_R \\
x &\leq x_M + \tfrac{1}{4} x_N - \tfrac{5}{4} x_P + \tfrac{25}{18} x_S \\
x &\leq x_M + \tfrac{1}{4} x_N - \tfrac{5}{4} x_P + \tfrac{25}{18} x_T \\
x &\leq x_M + \tfrac{1}{4} x_N - \tfrac{5}{4} x_P + \tfrac{25}{18} x_U
\end{aligned}
\qquad
\begin{aligned}
1 &\leq x_M \leq 1 \\
\tfrac{3}{10} x_M &\leq x_N \leq \tfrac{2}{5} x_M \\
\tfrac{1}{2} x_N &\leq x_O \leq \tfrac{3}{5} x_N \\
\tfrac{9}{10} x_O &\leq x_R \leq x_O \\
\tfrac{9}{10} x_O &\leq x_Q \leq x_O \\
\tfrac{4}{5} x_O &\leq x_P \leq \tfrac{9}{10} x_O \\
\tfrac{4}{5} x_P &\leq x_S \leq \tfrac{9}{10} x_P \\
\tfrac{4}{5} x_P &\leq x_T \leq \tfrac{9}{10} x_P \\
\tfrac{4}{5} x_P &\leq x_U \leq \tfrac{9}{10} x_P
\end{aligned}
$$

Figure 6: Generated Linear Inequalities in the Chaining Example

Thus, in this example, the tight upper answer is computed by solving a linear program that has 10 variables and 72 linear inequalities. Note that computing the tight upper answer by the classical linear programming approach would result in solving a linear program that has $2^9 = 512$ variables and $4 \cdot 9 - 2 = 34$ linear inequalities (see Section 4.6).

Let us now focus on the technical details. We subsequently generalize the function $H_1^\alpha$ of Section 4.1.1 in a straightforward way to compute greatest lower bounds in conditional constraint trees. Moreover, we present a linear program for computing the requested least upper bound in conditional constraint trees.

Let $Pr_1(C|B)$ denote $u_1$ for all $(C|B)[u_1, u_2] \in KB$. In the sequel, let $B$ be a node and let $C = B^\uparrow$. Again, the cases $C = B$, $C = B_1$ with a node $B_1 \neq B$, and $C = B_1 B_2 \ldots B_k$ with $k > 1$ nodes $B_1, B_2, \ldots, B_k$ refer to LEAF, CHAINING, and FUSION, respectively.

We define the generalized function $H_1^\alpha$ for computing greatest lower bounds in conditional constraint trees: let $H_1^\alpha(B, C) = \alpha_1$ (note that $\alpha_1$ will coincide with the greatest lower bound of $Pr(BL(C)) / Pr(B)$ subject to $Pr \models KB$ and $Pr(B) > 0$), where $\alpha_1$ in LEAF ($C = B$), CHAINING ($C = B_1$), and FUSION ($C = B_1 B_2 \ldots B_k$ with $k > 1$) is given by:

LEAF:

$$\alpha_1 = 1$$

CHAINING:

$$\alpha_1 = \max(0, Pr_1(C|B) \cdot (1 + \tfrac{H_1^\alpha(C, C^\uparrow) - 1}{Pr_1(B|C)}))$$

FUSION:

$$\alpha_1 = \max(0, 1 - k + \sum_{i=1}^{k} H_1^\alpha(B, B_i))$$

$H_1^\alpha$ is sound and globally complete with respect to $B$ and $C$ iff $H_1^\alpha(B, C) = \alpha_1$ is the greatest lower bound of $Pr(BL(C)) / Pr(B)$ subject to $Pr \in Mo(B, C)$ and $Pr(B) > 0$. Thus, the next theorem shows soundness and global completeness of $H_1^\alpha$.

**Theorem 4.3**

a) *For all probabilistic interpretations $Pr \in Mo(B, C)$, it holds $\alpha_1 \cdot Pr(B) \leq Pr(BL(C))$.*

b) *There exists a probabilistic interpretation $Pr \in Mo(B, C)$ with $Pr(B) > 0$, $\alpha_1 \cdot Pr(B) = Pr(BL(C))$, and $Pr(\overline{B} L(C)) = 0$ iff $L(C) \Rightarrow B$.*

**Proof.** The claims follow from Theorem 4.1. □

Next, we focus on the requested least upper bound, which is computed by solving a linear program as described in the two examples.

We start by defining the functions $I^\alpha$, $I^\beta$, and $I^\gamma$ over the variables $x_B$ ($B \in \mathcal{B}$). Let $I^\alpha(B, C) = \alpha_2$, $I^\beta(B, C) = \beta_2$, and $I^\gamma(B, C) = \gamma_2$, where $\alpha_2$, $\beta_2$, and $\gamma_2$ in LEAF ($C = B$), CHAINING ($C = B_1$), and FUSION ($C = B_1 B_2 \ldots B_k$ with $k > 1$) are given as follows:

LEAF:

$$\alpha_2 = x_B$$
$$\beta_2 = 0$$
$$\gamma_2 = x_B$$

Chaining:

$$\alpha_2 \;=\; \min\!\left(x_B, \frac{I^\gamma(C, C^\uparrow)}{Pr_1(B|C)}, \; x_C + \frac{I^\beta(C, C^\uparrow)}{Pr_1(B|C)}, \; x_B - x_C + \frac{I^\alpha(C, C^\uparrow)}{Pr_1(B|C)}\right)$$

$$\beta_2 \;=\; \min\!\left(\frac{1 - Pr_1(B|C)}{Pr_1(B|C)} \cdot x_C + \frac{I^\beta(C, C^\uparrow)}{Pr_1(B|C)}, \; \frac{I^\gamma(C, C^\uparrow)}{Pr_1(B|C)}\right)$$

$$\gamma_2 \;=\; \frac{I^\gamma(C, C^\uparrow)}{Pr_1(B|C)}$$

Fusion:

$$\alpha_2 \;=\; \min_{i \in [1:k]} I^\alpha(B, B_i)$$

$$\beta_2 \;=\; \min_{i \in [1:k]} I^\beta(B, B_i)$$

$$\gamma_2 \;=\; \min\!\left(\min_{i \in [1:k]} I^\gamma(B, B_i), \; \min_{i,j \in [1:k], i \neq j} (I^\alpha(B, B_i) + I^\beta(B, B_j))\right)$$

The system of linear inequalities $J(B, C)$ is defined as the least set of linear inequalities over $x_G \geq 0$ ($G \in \mathcal{B}(B, C)$) that contains $1 \leq x_B \leq 1$ and $u_1 \cdot x_G \leq x_H \leq u_2 \cdot x_G$ for all $(H|G)[u_1, u_2] \in KB(B, C)$ with $G \to H$ (that is, $G$ is the parent of $H$).

The intuition behind these definitions can now be described as follows.

Each $x_G$ ($G \in \mathcal{B}(B, C)$) that satisfies $J(B, C)$ corresponds to the exact conditional constraint tree $(\mathcal{B}(B, C), KB'(B, C))$, where $KB'(B, C)$ contains the pair $(H|G)[x_H/x_G, x_H/x_G]$ and $(G|H)[v_1, v_1]$ for each pair $(H|G)[u_1, u_2], (G|H)[v_1, v_2] \in KB(B, C)$ with $G \to H$.

We will show that the least upper bound of $Pr(BL(C))/Pr(B)$, $Pr(\overline{B}L(C))/Pr(B)$, and $Pr(L(C))/Pr(B)$ subject to $Pr \models KB'(B, C)$ and $Pr(B) > 0$ is given by $I^\alpha(B, C)$, $I^\beta(B, C)$, and $I^\gamma(B, C)$, respectively. It will then follow that the least upper bound of $Pr(BL(C))/Pr(B)$, $Pr(\overline{B}L(C))/Pr(B)$, and $Pr(L(C))/Pr(B)$ subject to $Pr \models KB(B, C)$ and $Pr(B) > 0$ is given by the maximum of $I^\alpha(B, C)$, $I^\beta(B, C)$, and $I^\gamma(B, C)$, respectively, subject to all $x_G$ ($G \in \mathcal{B}(B, C)$) satisfying $J(B, C)$.

That is, we implicitly performed the variable transformation described in the two examples. This transformation is indeed correct for conditional constraint trees:

**Lemma 4.4**

*a) If $x_G$ ($G \in \mathcal{B}(B, C)$) satisfies $J(B, C)$, then for all conditional constraints $(H|G)[u_1, u_2] \in KB(B, C)$ such that $G \to H$, there exists $u_H \in [u_1, u_2]$ with $x_H = u_H \cdot x_G$.*

*b) Let $u_H \in [u_1, u_2]$ for all $(H|G)[u_1, u_2] \in KB(B, C)$ such that $G \to H$. There exists $x_G$ ($G \in \mathcal{B}(B, C)$) with $J(B, C)$ and $x_H = u_H \cdot x_G$ for all nodes $H$ with parent $G$.*

**Proof.** a) For all nodes $H$ with parent $G$, let $u_H$ be defined by $u_H = x_H / x_G$.

b) Let $x_B = 1$, and for all nodes $H$ with parent $G$, let $x_H$ be defined by $x_H = u_H \cdot x_G$. $\square$

We are now ready to formulate an optimization problem for computing the requested least upper bound.

**Theorem 4.5** *Let $X_2$ be the maximum of $x$ subject to $x \leq I^\alpha(E, E^\uparrow)$ and $J(E, E^\uparrow)$.*

*a) $Pr(EL(E^\uparrow)) \leq X_2 \cdot Pr(E)$ for all $Pr \in Mo(E, E^\uparrow)$.*

*b) There exists $Pr \in Mo(E, E^\uparrow)$ with $Pr(E) > 0$ and $Pr(EL(E^\uparrow)) = X_2 \cdot Pr(E)$.*

**Proof.** Let $Pr(B|C) = v_1$ for all $(B|C)[v_1, v_2] \in KB$ such that $B \to C$. By Theorem 4.2, the requested least upper bound is the maximum of $x$ subject to $x \leq H_J^\alpha(E, E^\uparrow)$ and $Pr(C|B) = u_C \in [u_1, u_2]$ for all $(C|B)[u_1, u_2] \in KB$ such that $B \to C$. By Lemma 4.4, we can equivalently maximize $x$ subject to $x \leq I^\alpha(E, E^\uparrow)$ and $J(E, E^\uparrow)$. $\square$

We now wonder how to solve the generated optimization problem, since $I^\alpha(E, E^\uparrow)$ may still contain min-operations that cannot be tackled by linear programming. Moreover, given a method for solving this optimization problem, we are also interested in a rough idea on the overall time complexity of computing the requested least upper bound this way. Finally, we are interested in possible improvements to increase efficiency. These topics are discussed in the rest of this section.

If $I^\alpha(E, E^\uparrow)$ does not contain any min-operations at all, then the generated optimization problem is already a linear program. Otherwise, it can easily be transformed into a linear program. In a first transformation step, all inner min-operations are eliminated. This can easily be done due to the well-structuredness of $I^\alpha(E, E^\uparrow)$. In a second step, the only remaining outer min-operation is eliminated by introducing exactly one linear inequality for each contained operand. In these linear inequalities, the operands of the outer min-operation are upper bounds of $x$.

To get a rough idea on the time complexity of computing the requested least upper bound this way, we must analyze the size of the generated linear programs. It is given by the number of variables, the number of linear inequalities in $J(E, E^\uparrow)$, and the number of linear inequalities extracted from $x \leq I^\alpha(E, E^\uparrow)$. The latter is quite worrying, since $I^\gamma(B, C)$ in FUSION seems to produce many min-operands. Moreover, $I^\gamma(B, C)$ in FUSION contains $I^\alpha(B, B_i)$, and $I^\alpha(B, C)$ in CHAINING contains $I^\gamma(C, C^\uparrow)$. So, due to this crossed dependency, the overall number of generated linear inequalities is likely to 'explode' for trees that branch very often.

To avoid these problems, we introduce the auxiliary functions $J^\alpha$, $J^\beta$, and $J^\gamma$ over the variables $x_B$ ($B \in \mathcal{B}$). Let $J^\alpha(B, C) = \alpha_2'$, $J^\beta(B, C) = \beta_2'$, and $J^\gamma(B, C) = \gamma_2'$, where $\alpha_2'$, $\beta_2'$, and $\gamma_2'$ in LEAF ($C = B$), CHAINING ($C = B_1$), and FUSION ($C = B_1 B_2 \ldots B_k$ with $k > 1$) are given as follows:

LEAF:

$$\alpha_2' = x_B$$
$$\beta_2' = 0$$
$$\gamma_2' = x_B$$

CHAINING:

$$\alpha_2' = \min(x_B, \ x_C + \frac{J^\beta(C, C^\uparrow)}{Pr_1(B|C)}, \ x_B - x_C + \frac{J^\alpha(C, C^\uparrow)}{Pr_1(B|C)})$$
$$\beta_2' = \frac{1 - Pr_1(B|C)}{Pr_1(B|C)} \cdot x_C + \frac{J^\beta(C, C^\uparrow)}{Pr_1(B|C)}$$
$$\gamma_2' = \frac{J^\gamma(C, C^\uparrow)}{Pr_1(B|C)}$$

218

FUSION:

$$\alpha_2' = \min_{i \in [1:k]} J^\alpha(B, B_i)$$

$$\beta_2' = \min_{i \in [1:k]} J^\beta(B, B_i)$$

$$\gamma_2' = \min(\min_{i \in [1:k]} J^\gamma(B, B_i), \min_{i,j \in [1:k], i \neq j} (J^\alpha(B, B_i) + J^\beta(B, B_j)))$$

Note that $\alpha_2'$ in CHAINING can be separated into the cases $C^\uparrow = C$ and $C^\uparrow \neq C$. Since simply $\alpha_2' = x_C$ for $C^\uparrow = C$, we reduce the number of generated linear inequalities this way.

The next lemma shows that the functions $I^\alpha$, $I^\beta$, and $I^\gamma$ can be expressed in terms of the auxiliary functions $J^\alpha$, $J^\beta$, and $J^\gamma$.

**Lemma 4.6** *For all $x_B$ ($B \in \mathcal{B}$) that satisfy $J(E, E^\uparrow)$:*

$$\alpha_2 = \min(\alpha_2', \gamma_2'), \ \beta_2 = \min(\beta_2', \gamma_2'), \ and \ \gamma_2 = \gamma_2' \,.$$

**Proof sketch.** The claim can be proved by induction on the recursive definition of the functions $I^\alpha$, $I^\beta$, and $I^\gamma$. $\square$

Briefly, by Theorem 4.3, Theorem 4.5, and Lemma 4.6, the tight answer to the premise-restricted complete query $\exists (F|E)[x_1, x_2]$ is given by $\{x_1/H_1^\alpha(E, E^\uparrow), x_2/X_2\}$, where $X_2$ is the maximum of $x$ subject to $x \leq J^\alpha(E, E^\uparrow)$, $x \leq J^\gamma(E, E^\uparrow)$, and $J(E, E^\uparrow)$.

In our example, we get $\{x_1/0.00, x_2/0.27\}$ as the tight answer to the premise-restricted complete query $\exists (\mathsf{QRSTU}|\mathsf{M})[x_1, x_2]$ to the conditional constraint tree in Fig. 2, right side.

The time complexity of computing the requested greatest lower bound and especially the requested least upper bound this way is analyzed in Section 4.5.

## 4.2 Strongly Conclusion-Restricted Complete Queries

We now focus on computing the tight answer to strongly conclusion-restricted complete queries to general conditional constraint trees. In the sequel, let $(\mathcal{B}, KB)$ be a conditional constraint tree and let $\exists (F|E)[x_1, x_2]$ be a strongly conclusion-restricted complete query.

The tight upper answer to $\exists (F|E)[x_1, x_2]$ is always given by $\{x_2/1\}$. To compute the tight lower answer to $\exists (F|E)[x_1, x_2]$, we first compute the tight lower answer $\{y_1/u_1\}$ to the premise-restricted complete query $\exists (E|F)[y_1, y_2]$. We then distinguish the following cases:

If $u_1 > 0$, then the tight lower answer to $\exists (F|E)[x_1, x_2]$ is computed locally by a function $H_1^\delta$ (like the tight lower answer to premise-restricted complete queries in Section 4.1.2). If $u_1 = 0$ and $E \Rightarrow F$, then the tight lower answer to $\exists (F|E)[x_1, x_2]$ is given by $\{x_1/1\}$. Otherwise, the tight lower answer to $\exists (F|E)[x_1, x_2]$ is given by $\{x_1/0\}$.

We now focus on the technical details. Let $(\mathcal{B}, \rightarrow)$ be the directed graph that belongs to the premise-restricted complete query $\exists (E|F)[y_1, y_2]$ (see Section 4.1.1). Let $Pr_1(B|C)$ denote $v_1$ for all $(B|C)[v_1, v_2] \in KB$. In the sequel, let $B$ be a node and let $C = B^\uparrow$. Again, the cases $C = B$, $C = B_1$ with a node $B_1 \neq B$, and $C = B_1 B_2 \ldots B_k$ with $k > 1$ nodes $B_1, B_2, \ldots, B_k$ refer to LEAF, CHAINING, and FUSION, respectively.

We define the function $H_1^\delta$ for computing greatest lower bounds in the case $H_1^\alpha(B, C) > 0$ as follows. Let $H_1^\delta(B, C) = \delta_1$ (note that $\delta_1$ will coincide with the greatest lower bound of $Pr(BL(C))/Pr(L(C))$ subject to $Pr \models KB$ and $Pr(L(C)) > 0$), where $\delta_1$ in LEAF $(C = B)$, CHAINING $(C = B_1)$, and FUSION $(C = B_1 B_2 \ldots B_k$ with $k > 1)$ is given as follows (note that $H_1^\alpha(C, C^\uparrow)$ and $H_1^\alpha(B, B_i)$ are defined like in Section 4.1.2):

LEAF:

$$\delta_1 \quad = \quad 1$$

CHAINING:

$$\delta_1 \quad = \quad H_1^\delta(C, C^\uparrow) \cdot (1 + \tfrac{Pr_1(B|C) - 1}{H_1^\alpha(C, C^\uparrow)})$$

FUSION:

$$\delta_1 \quad = \quad \left( 1 + \frac{\min\limits_{i \in [1:k]} H_1^\alpha(B, B_i) \cdot (1/H_1^\delta(B, B_i) - 1)}{1 - k + \sum\limits_{i=1}^{k} H_1^\alpha(B, B_i)} \right)^{-1}$$

By induction on the definition of $H_1^\delta$, it is easy to see that $H_1^\alpha(B, C) > 0$ entails that $\delta_1$ is defined and that $\delta_1 > 0$ (note that $H_1^\alpha(B, C) = \alpha_1$ in LEAF, CHAINING, and FUSION is defined like in Section 4.1.2). In this case, $H_1^\delta$ is sound and globally complete with respect to $B$ and $C$ iff $H_1^\delta(B, C) = \delta_1$ is the greatest lower bound of $Pr(BL(C)) / Pr(L(C))$ subject to $Pr \in Mo(B, C)$ and $Pr(L(C)) > 0$. Thus, the next theorem shows soundness and global completeness of $H_1^\delta$. It also shows that, for $C = B_1 B_2 \ldots B_k$ with $k > 1$, the least upper bound of $Pr(BL(C)) / Pr(L(C))$ subject to $Pr \in Mo(B, C)$ and $Pr(L(C)) > 0$ is given by 1.

**Theorem 4.7**

*a) If $\alpha_1 > 0$, then for all $Pr \in Mo(B, C)$, it holds $\delta_1 \cdot Pr(L(C)) \leq Pr(BL(C))$.*

*b) If $\alpha_1 > 0$, then there is a probabilistic interpretation $Pr \in Mo(B, C)$ with $Pr(B) > 0$, $Pr(L(C)) > 0$, $\delta_1 \cdot Pr(L(C)) = Pr(BL(C))$, and $\alpha_1 \cdot Pr(B) = Pr(BL(C))$.*

*c) If $\alpha_1 > 0$ and $C = B_1 B_2 \ldots B_k$ with $k > 1$, then there is some $Pr \in Mo(B, C)$ with $Pr(B) > 0$, $Pr(L(C)) > 0$, $1 \cdot Pr(L(C)) = Pr(BL(C))$, and $\alpha_1 \cdot Pr(B) = Pr(BL(C))$.*

*d) If $\alpha_1 = 0$ and $C = B_1 B_2 \ldots B_k$ with $k > 1$, then for each $\varepsilon > 0$ there is some $Pr \in Mo(B, C)$ with $Pr(B) > 0$, $Pr(L(C)) > 0$, $1 \cdot Pr(L(C)) = Pr(BL(C))$, and $\varepsilon \cdot Pr(B) \geq Pr(BL(C))$.*

**Proof.** The proof is given in full detail in Appendix C. $\square$

We are now ready to give the following characterization of tight answers to strongly conclusion-restricted complete queries to conditional constraint trees.

**Theorem 4.8** *Let $(\mathcal{B}, KB)$ be a conditional constraint tree and let $\exists(F|E)[x_1, x_2]$ be a strongly conclusion-restricted complete query. Let the tight lower answer to the premise-restricted complete query $\exists(E|F)[y_1, y_2]$ be given by $\{y_1/u_1\}$.*

(1) *If $u_1 > 0$, then the tight answer to $\exists(F|E)[x_1, x_2]$ is given by $\{x_1/H_1^\delta(F, F^\uparrow), x_2/1\}$.*

(2) *If $u_1 = 0$ and $E \Rightarrow F$, then the tight answer to $\exists(F|E)[x_1, x_2]$ is given by $\{x_1/1, x_2/1\}$.*

(3) *Otherwise, the tight answer to $\exists(F|E)[x_1, x_2]$ is given by $\{x_1/0, x_2/1\}$.*

**Proof.** The proof is given in full detail in Appendix C. $\square$

### 4.3 Complete Queries

We now show that the problem of computing tight answers to complete queries can be reduced to the problems of computing tight answers to premise-restricted complete queries and of computing tight answers to strongly conclusion-restricted complete queries.

In detail, a complete query is premise-restricted, it is strongly conclusion-restricted, or it can be reduced to premise-restricted complete queries and to strongly conclusion-restricted complete queries. For example, given the complete query $\exists(\mathsf{STU}|\mathsf{MQR})[x_1, x_2]$ to the conditional constraint tree in Fig. 2, right side, we first compute the tight answer $\{y_1/u_1,\ y_2/u_2\}$ to the premise-restricted complete query $\exists(\mathsf{MQR}|\mathsf{O})[y_1, y_2]$ (directed to the corresponding subtree) and the tight answer $\{z_1/v_1,\ z_2/v_2\}$ to the strongly conclusion-restricted complete query $\exists(\mathsf{O}|\mathsf{MQR})[z_1, z_2]$ (directed to the corresponding subtree). We then generate a new conditional constraint tree by replacing the subtree over the nodes $\mathsf{M}$, $\mathsf{N}$, $\mathsf{O}$, $\mathsf{Q}$, and $\mathsf{R}$ by the pair of conditional constraints $(\mathsf{B}|\mathsf{O})[u_1, u_2]$ and $(\mathsf{O}|\mathsf{B})[v_1, v_2]$ over the nodes $\mathsf{B}$ and $\mathsf{O}$ (note that $\mathsf{B}$ represents $\mathsf{MQR}$). Finally, we compute the tight answer to the premise-restricted complete query $\exists(\mathsf{STU}|\mathsf{B})[x_1, x_2]$ to the new conditional constraint tree.

Note that this reduction can always be done, since for each query $\exists(F|E)[x_1, x_2]$, all paths from a basic event in $E$ to a basic event in $F$ have at least one basic event in common.

**Theorem 4.9** *Let $(\mathcal{B}, KB)$ be a conditional constraint tree and let $\exists(F|E)[x_1, x_2]$ be a complete query that is not premise-restricted and not strongly conclusion-restricted.*

*a) There exists a basic event $G \in \mathcal{B}$ and two conditional constraint trees $(\mathcal{B}_1, KB_1)$ and $(\mathcal{B}_2, KB_2)$ such that $\mathcal{B}_1 \cap \mathcal{B}_2 = \{G\}$, $\mathcal{B}_1 \cup \mathcal{B}_2 = \mathcal{B}$, and $\exists(G|E)[z_1, z_2]$ is a strongly conclusion-restricted complete query to $(\mathcal{B}_1, KB_1)$.*

*b) Let the tight answer to the premise-restricted complete query $\exists(E|G)[y_1, y_2]$ to $(\mathcal{B}_1, KB_1)$ be given by $\{y_1/u_1,\ y_2/u_2\}$ and let the tight answer to the strongly conclusion-restricted complete query $\exists(G|E)[z_1, z_2]$ to $(\mathcal{B}_1, KB_1)$ be given by $\{z_1/v_1,\ z_2/v_2\}$.*

*(1) If $u_1 > 0$, then also $v_1 > 0$ and the tight answer to the complete query $\exists(F|E)[x_1, x_2]$ to $(\mathcal{B}, KB)$ coincides with the tight answer to the premise-restricted complete query $\exists(F|B)[x_1, x_2]$ to $(\mathcal{B}_2 \cup \{B\}, KB_2 \cup \{(B|G)[u_1, u_2], (G|B)[v_1, v_2]\})$, where $B$ is a new basic event with $B \notin \mathcal{B}_2$. In particular, for exact conditional constraint trees $(\mathcal{B}, KB)$, the tight answer to the complete query $\exists(F|E)[x_1, x_2]$ is given by:*

$$\{x_1/\max(0, v_1 - \tfrac{v_1}{u_1} + \tfrac{v_1 s_1}{u_1}),\ x_2/\min(1, 1 - v_1 + \tfrac{v_1 s_2}{u_1}, \tfrac{t_2}{t_2 - s_2 + u_1})\}\,,$$

*where $s_1 = H_1^{\alpha}(G, G^{\uparrow})$, $s_2 = H_2^{\alpha}(G, G^{\uparrow})$, and $t_2 = H_2^{\gamma}(G, G^{\uparrow})$ (note that $H_1^{\alpha}$, $H_2^{\alpha}$, and $H_2^{\gamma}$ are defined like in Section 4.1.1).*

*(2) If $u_1 = 0$, $v_1 = 1$, and $G \Rightarrow F$, then the tight answer to the complete query $\exists(F|E)[x_1, x_2]$ to $(\mathcal{B}, KB)$ is given by $\{x_1/1,\ x_2/1\}$.*

*(3) Otherwise, the tight answer to the complete query $\exists(F|E)[x_1, x_2]$ to $(\mathcal{B}, KB)$ is given by $\{x_1/0,\ x_2/1\}$.*

**Proof.** The proof is given in full detail in Appendix D. □

#### 4.4 Queries

The problem of computing tight answers to queries can be reduced to the more specialized problem of calculating tight answers to complete queries.

More precisely, given a query $\exists (F|E)[x_1, x_2]$ to a conditional constraint tree $(\mathcal{B}, KB)$, a complete query $\exists (F'|E')[x_1, x_2]$ to a conditional constraint tree $(\mathcal{B}', KB')$ is generated by:

1. While $(\mathcal{B}, KB)$ contains a leaf $B$ that is not contained in $EF$: remove $B$ from $\mathcal{B}$ and remove the corresponding pair $(C|B)[u_1, u_2], (B|C)[v_1, v_2] \in KB$ from $KB$.

2. While $EF$ contains a basic event $B$ that is not a leaf in $(\mathcal{B}, KB)$: increase $\mathcal{B}$ by a new basic event $B'$, increase $KB$ by the pair $(B'|B)[1, 1]$ and $(B|B')[1, 1]$, and replace each occurrence of $B$ in $\exists (F|E)[x_1, x_2]$ by the new basic event $B'$.

It remains to show that the generated probabilistic deduction problem has the same solution as the original probabilistic deduction problem:

**Theorem 4.10** *The tight answer to the query $\exists (F|E)[x_1, x_2]$ to $(\mathcal{B}, KB)$ coincides with the tight answer to the complete query $\exists (F'|E')[x_1, x_2]$ to $(\mathcal{B}', KB')$.*

**Proof.** Let $(\mathcal{B}'', KB'')$ be the conditional constraint tree that is generated in step 1 and let $(F|E)[u_1, u_2]$ be a tight logical consequence of $KB''$. We now show that $(F|E)[u_1, u_2]$ is also a tight logical consequence of $KB$. First, $(F|E)[u_1, u_2]$ is a logical consequence of $KB$, since $KB''$ is a subset of $KB$. Moreover, each model $Pr''$ of $KB''$ (that is defined on all atomic events over $\mathcal{B}''$) can be extended to a model $Pr$ of $KB$ (that is defined on all atomic events over $\mathcal{B}$) with $Pr(A) = s \cdot Pr''(A)$ for all atomic events $A$ over $\mathcal{B}''$ that are different from the conjunction of all negated basic events in $\mathcal{B}''$, where $s$ is a real number from $(0, 1]$. This model can be constructed inductively like in the proof of Theorem 3.2. Thus, for $u \in [u_1, u_2]$, $Pr''(E) > 0$ and $u \cdot Pr''(E) = Pr''(EF)$ entails $Pr(E) > 0$ and $u \cdot Pr(E) = Pr(EF)$.

Finally, $(F|E)[u_1, u_2]$ is a tight logical consequence of $KB''$ iff $(F'|E')[u_1, u_2]$ is a tight logical consequence of $KB'$, since we just introduce synonyms for basic events in step 2. $\square$

#### 4.5 Computational Complexity

##### 4.5.1 EXACT CONDITIONAL CONSTRAINT TREES

We now show that for exact conditional constraint trees, our technique to compute the tight answer to queries runs in linear time in the number of nodes of the tree. In the sequel, let $(\mathcal{B}, KB)$ be an exact conditional constraint tree and let $n$ denote its number of nodes.

**Lemma 4.11** *The tight answer to a premise-restricted or strongly conclusion-restricted complete query can be computed in linear time in $n$.*

**Proof.** For exact conditional constraint trees, our approach to compute the tight upper answer to premise-restricted complete queries by $H_2^\alpha$, $H_2^\beta$, and $H_2^\gamma$ runs in time $O(n)$:

The directed tree can be computed in time $O(n)$. An initialization of a leaf with a constant number of assignments is performed exactly for each leaf of the directed tree, a chaining with a constant number of arithmetic operations is performed exactly for each arrow of the directed tree. Hence, initializing all leaves and performing all chainings runs

222

in time $O(n)$. A fusion is done for each branching of the directed tree, using linear time in the number of branches. Thus, all fusions together run in time $O(n)$.

Even for general conditional constraint trees, the tight lower answer to premise-restricted complete queries, and hence also the tight answer to strongly conclusion-restricted complete queries, is analogously computed in time $O(n)$. $\square$

**Theorem 4.12** *The tight answer to a query can be computed in linear time in $n$.*

**Proof.** We assume that the set of basic events $\mathcal{B}$ is totally ordered and that the basic events in the conjunctive events $E$ and $F$ of the query $\exists (F|E)[x_1, x_2]$ are written in this order.

First, the query is reduced to a complete query according to Section 4.4. This reduction can be done in time $O(n)$. Now, if the generated complete query is premise-restricted or strongly conclusion-restricted, then the claim follows immediately from Lemma 4.11.

Otherwise, the generated complete query is reduced to premise-restricted and strongly conclusion-restricted complete queries according to Section 4.3. Also this reduction can be done in time $O(n)$, since the basic event $G$ in Theorem 4.9 a) is computable in time $O(n)$. Hence, the claim follows from Theorem 4.9 and Lemma 4.11. Note that $t_2 = H_2^\gamma(G, G^\uparrow)$ in Theorem 4.9 b) (1) can also be computed in time $O(n)$. $\square$

### 4.5.2 Conditional Constraint Trees

For general conditional constraint trees, our technique to compute the tight lower answer to queries runs still in linear time, while our technique to compute the tight upper answer to queries runs in polynomial time in the number of nodes of the tree. In the sequel, let $(\mathcal{B}, KB)$ be a general conditional constraint tree and let $n$ denote its number of nodes.

**Lemma 4.13**

*a) The tight lower answer to a premise-restricted complete query and the tight answer to a strongly conclusion-restricted complete query can be computed in linear time in $n$.*

*b) The tight upper answer to a premise-restricted complete query can be computed in polynomial time in $n$.*

**Proof.** a) The claim is already shown in the proof of Lemma 4.11.

b) Our linear programming technique to compute the tight upper answer to premise-restricted complete queries runs in polynomial time in $n$:

Linear programming runs in polynomial time in the size of the linear programs (Papadimitriou & Steiglitz, 1982; Schrijver, 1986), where the size of a linear program is given by its number of variables and its number of linear inequalities.

We now show that the size of our linear programs in Section 4.1.2 is polynomial in $n$. The number of variables is $n + 1$. The number of linear inequalities in $J(E, E^\uparrow)$ is $2n$. By induction on the recursive definition of $J^\alpha$, $J^\beta$, and $J^\gamma$, it can be shown that the number of min-operands in $J^\alpha(B, C)$, $J^\beta(B, C)$, and $J^\gamma(B, C)$ is limited by $|\mathcal{B}(B, C)|^2$, $|\mathcal{B}(B, C)|$, and $|\mathcal{B}(B, C)|^4$, respectively. Hence, the number of linear inequalities extracted from $x \leq J^\alpha(E, E^\uparrow)$ and $x \leq J^\gamma(E, E^\uparrow)$ is limited by $|\mathcal{B}(E, E^\uparrow)|^2 + |\mathcal{B}(E, E^\uparrow)|^4 = n^2 + n^4$. Thus, the overall number of generated linear inequalities $l$ is limited by $l_u = 2n + n^2 + n^4$.

Finally, note that $l_u$ is a very rough upper bound for $l$, in many conditional constraint trees (especially in those that branch very rarely), $l$ is much lower than $l_u$. For example, taking a complete binary tree with $n = 127$ nodes, we get only $l = 19\,964$ compared to $l_u = 260\,161\,024$. In the example of Section 4.1.2 with $n = 9$ nodes, we get only $l = 72$ compared to $l_u = 6\,660$. Another example is a tree that is degenerated to a chain of basic events. In this case, we even get $l = 5n + 1$, that is, the overall number of generated linear inequalities is linear in $n$. $\square$

**Theorem 4.14**

*a) The tight lower answer to a query can be computed in linear time in $n$.*

*b) The tight upper answer to a query can be computed in polynomial time in $n$.*

**Proof.** We assume that the set of basic events $\mathcal{B}$ is totally ordered and that the basic events in the conjunctive events $E$ and $F$ of the query $\exists(F|E)[x_1, x_2]$ are written in this order.

Like in the proof of Theorem 4.12, the query is reduced to a complete query according to Section 4.4. This reduction can be done in time $O(n)$. Now, if the generated complete query is premise-restricted or strongly conclusion-restricted, then the claims follow immediately from Lemma 4.13.

Otherwise, the generated complete query is reduced to premise-restricted and strongly conclusion-restricted complete queries according to Section 4.3. Again, this reduction can be done in time $O(n)$, since the basic event $G$ in Theorem 4.9 a) is computable in time $O(n)$. Thus, the claims follow from Theorem 4.9 and Lemma 4.13. Note that in Theorem 4.9 b) (1), the tight lower answer to $\exists(F|B)[x_1, x_2]$ can be computed without $u_2$ and $v_2$. $\square$

### 4.6 Comparison with the Classical Linear Programming Approach

As a comparison, we now briefly describe how probabilistic deduction in conditional constraint trees can be done by the classical linear programming approach (Paaß, 1988; van der Gaag, 1991; Amarger et al. 1991; Hansen et al. 1995). In the sequel, let $\exists(F|E)[x_1, x_2]$ be a query to an exact or general conditional constraint tree $(\mathcal{B}, KB)$ over $n$ nodes.

The tight answer to $\exists(F|E)[x_1, x_2]$ can be computed by solving two linear programs. In detail, the requested greatest lower and least upper bound are given by the optimal values of the following two linear programs with $x_A \geq 0$ ($A \in \mathcal{A}_\mathcal{B}$) and opt $\in \{\min, \max\}$:

$$\text{opt } \sum_{A \in \mathcal{A}_\mathcal{B},\, A \Rightarrow EF} x_A \text{ subject to}$$

$$\sum_{A \in \mathcal{A}_\mathcal{B},\, A \Rightarrow E} x_A \;=\; 1$$

$$\sum_{A \in \mathcal{A}_\mathcal{B},\, A \Rightarrow GH} x_A \;\geq\; u_1 \cdot \sum_{A \in \mathcal{A}_\mathcal{B},\, A \Rightarrow G} x_A \text{ for all } (H|G)[u_1, u_2] \in KB$$

$$\sum_{A \in \mathcal{A}_\mathcal{B},\, A \Rightarrow GH} x_A \;\leq\; u_2 \cdot \sum_{A \in \mathcal{A}_\mathcal{B},\, A \Rightarrow G} x_A \text{ for all } (H|G)[u_1, u_2] \in KB$$

That is, the tight answer is computed by solving two linear programs with $2^n$ variables and $4n - 2$ linear inequalities. For example, the tight answer to the premise-restricted complete query $\exists(\mathsf{QRSTU}|\mathsf{M})[x_1, x_2]$ to the conditional constraint trees in Fig. 2 yields two linear programs with $2^9 = 512$ variables and $4 \cdot 9 - 2 = 34$ linear inequalities.

Hence, if we now solve these two linear programs by the standard simplex method or the standard interior-point technique, then we need immediately exponential time in $n$. It is still an open question whether column generation techniques can help to solve the two linear programs in less than exponential time in $n$ in the worst case.

## 5. Comparison with Bayesian Networks

In this section, we briefly discuss the relationship between conditional constraint trees and Bayesian networks (Pearl, 1988).

A Bayesian network is defined by a directed acyclic graph $G$ over discrete random variables $X_1, X_2, \ldots, X_n$ as nodes and by a conditional probability distribution $Pr(X_i|\mathbf{pa}(X_i))$ for each random variable $X_i$ and each instantiation $\mathbf{pa}(X_i)$ of its parents $pa(X_i)$. It specifies a unique joint probability distribution $Pr$ over $X_1, X_2, \ldots, X_n$ by:

$$Pr(X_1, X_2, \ldots, X_n) = \prod_{i=1}^{n} Pr(X_i|pa(X_i)).$$

That is, the joint probability distribution $Pr$ is uniquely determined by the conditional distributions $Pr(X_i|\mathbf{pa}(X_i))$ and certain conditional independencies encoded in $G$.

Hence, Bayesian trees (that is, Bayesian networks that have a directed tree as associated directed acyclic graph) with only binary random variables seem to be very close to exact conditional constraint trees. However, exact and general conditional constraint trees are associated with an undirected tree that does not encode any independencies! For this reason, exact and general conditional constraint trees describe convex sets of joint probability distributions rather than single joint probability distributions.

But, would it be possible to additionally assume certain independencies? Of course, with each exact or general conditional constraint tree $(\mathcal{B}, KB)$, we can associate all probabilistic interpretations $Pr$ that are models of $KB$ and that have additionally the undirected tree $(\mathcal{B}, \leftrightarrow)$ as an I-map (Pearl, 1988). That is, we would have independencies without causal directionality like in Markov trees (Pearl, 1988). However, this idea does not carry us to a single probabilistic interpretation (neither for exact conditional constraint trees, nor for general conditional constraint trees), and it is an interesting topic of future research to investigate how the computation of tight answers in exact and general conditional constraint trees changes under this kind of independencies (which yield tighter bounds, since they reduce the number of models of exact and general conditional constraint trees).

Finally, if we additionally fix the probability of exactly one node, then an exact conditional constraint tree under the described independencies specifies exactly one probabilistic interpretation (note that, to keep satisfiability, the probability of a node must respect certain upper bounds, which are entailed by the exact conditional constraint tree). But, such exact conditional constraint trees are in fact Bayesian trees with only binary random variables.

## 6. Summary and Conclusions

We showed that globally complete probabilistic deduction with conditional constraints over basic events is NP-hard. We then concentrated on the special case of probabilistic deduction in exact and general conditional constraint trees. We presented very efficient techniques for globally complete probabilistic deduction. More precisely, for exact conditional constraint trees, we presented a local approach that runs in linear time in the size of the conditional constraint trees. For general conditional constraint trees, we introduced a global approach that runs in polynomial time in the size of the conditional constraint trees.

Probabilistic deduction in conditional constraint trees is motivated by previous work in the literature on inference rules. It generalizes patterns of commonsense reasoning that have been thoroughly studied in this work. Hence, we presented a new class of tractable probabilistic deduction problems, which are driven by artificial intelligence applications.

It is also important to note that the deduction process in exact and general conditional constraint trees can easily be elucidated in a graphical way. For example, the computation of the tight answer to the premise-restricted complete query $\exists(\mathsf{QRSTU}|\mathsf{M})[x_1, x_2]$ to the exact conditional constraint tree in Fig. 2, left side, can be illustrated by labeling each node of the directed tree in Fig. 3 with the corresponding tightest bounds of Table 1.

Like Bayesian networks, conditional constraint trees are well-structured probabilistic knowledge bases that have an intuitive graphical representation. Differently from Bayesian networks, conditional constraint trees do not encode any probabilistic independencies. Thus, they can also be understood as a complement to Bayesian networks, useful for restricted applications in which well-structured independencies do not hold or are difficult to access.

Conditional constraint trees are quite restricted in their expressive power. However, in more general probabilistic knowledge bases, probabilistic deduction in conditional contraint trees may always act as local inference rules. For example, in case we desire explanatory information on some specific local deductions from a subset of the whole knowledge base (which could especially be useful in the design phase of a probabilistic knowledge base).

An important conclusion of this paper concerns the question whether to perform probabilistic deduction by the iterative application of inference rules or by linear programming. The techniques of this paper have been elaborated by following the idea of inference rules in probabilistic deduction. Hence, on the one hand, this paper shows that the idea of inference rules can indeed bring us to efficient techniques for globally complete probabilistic deduction in restricted settings. However, on the other hand, given the technical complexity of the corresponding proofs, it seems unlikely that these results can be extended to probabilistic knowledge bases that are significantly more general than conditional constraint trees.

That is, as far as significantly more general probabilistic deduction problems with conditional constraints are concerned, the iterative application of inference rules does not seem to be very promising for globally complete probabilistic deduction. Note that a similar conclusion is drawn in a companion paper (1998a, 1999a), which shows the limits of locally complete inference rules for probabilistic deduction under taxonomic knowledge.

For example, probabilistic deduction from probabilistic logic programs that do not assume probabilistic independencies (Ng & Subrahmanian 1993, 1994; Lukasiewicz, 1998d) should better not be done by the iterative application of inference rules. Note that much more promising techniques are, for example, global techniques by linear programming (Lukasiewicz, 1998d) and in particular approximation techniques based on truth-functional many-valued logics (Lukasiewicz 1998b, 1999b).

## Acknowledgements

I am very grateful to Michael Wellman and the referees for their useful comments. I also want to thank Thomas Eiter for valuable comments on an earlier version of this paper. This paper is an extended and revised version of a paper that appeared in *Principles of Knowledge Representation and Reasoning: Proceedings of the 6th International Conference*, pp. 380–391.

## Appendix A. Preliminaries of the Proofs for Sections 4.1 to 4.3

In this section, we make some technical preparations for the proofs of Theorems 4.1, 4.2, 4.7, 4.8, and 4.9. In the sequel, we use the notation

$$
\begin{array}{cc|c}
x_{1,1} & x_{1,2} & r_1 \\
x_{2,1} & x_{2,2} & r_2 \\
\hline
c_1 & c_2 &
\end{array}
$$

as an abbreviation of the following system of equations:

$$
\begin{aligned}
(11) \qquad x_{1,1} + x_{1,2} &= r_1 & x_{1,1} + x_{2,1} &= c_1 \\
x_{2,1} + x_{2,2} &= r_2 & x_{1,2} + x_{2,2} &= c_2 \; .
\end{aligned}
$$

The next lemma provides the optimal values of two linear programs to be solved in the proofs of Theorems 4.1, 4.2, 4.7, 4.8, and 4.9.

**Lemma A.1** *Let $r_1, r_2, c_1, c_2 \geq 0$ with $r_1 + r_2 = c_1 + c_2$. For all $i, j \in \{1, 2\}$:*

*a)* $\min(r_i, c_j) = \max x_{i,j}$ *subject to (11) and $x_{n,m} \geq 0$ for all $n, m \in \{1, 2\}$.*

*b)* $\max(0, r_i - c_{3-j}) = \min x_{i,j}$ *subject to (11) and $x_{n,m} \geq 0$ for all $n, m \in \{1, 2\}$.*

**Proof.** The claims can easily be verified (Lukasiewicz, 1996). □

Let us assume that a conditional constraint tree is the union of two subtrees that have just one node in common. A model of each subtree and a third model related to the common node can now be combined to a model of the whole conditional constraint tree. This important result follows from the next lemma.

**Lemma A.2** *Let $\mathcal{B}_1$ and $\mathcal{B}_2$ be sets of basic events with $\mathcal{B}_1 \cap \mathcal{B}_2 = \emptyset$. Let $B_0$ be a new basic event that is not contained in $\mathcal{B}_1 \cup \mathcal{B}_2$. Let $Pr_1$ and $Pr_2$ be probabilistic interpretations on the atomic events over $\mathcal{B}_1 \cup \{B_0\}$ and $\mathcal{B}_2 \cup \{B_0\}$, respectively. Let $B_1$ and $B_2$ be conjunctive events over $\mathcal{B}_1$ and $\mathcal{B}_2$, respectively. Let $Pr_0$ be a probabilistic interpretation on the atomic events over $\{B_0, B_1, B_2\}$ with $Pr_0(H_0 H_1) = Pr_1(H_0 H_1)$ and $Pr_0(H_0 H_2) = Pr_2(H_0 H_2)$ for all atomic events $H_0$, $H_1$, and $H_2$ over $\{B_0\}$, $\{B_1\}$, and $\{B_2\}$, respectively.*

*There is a probabilistic interpretation $Pr$ on the atomic events over $\mathcal{B}_1 \cup \mathcal{B}_2 \cup \{B_0\}$ with:*

$$
\begin{aligned}
(12) \qquad Pr(H_0 H_1 H_2) &= Pr_0(H_0 H_1 H_2), \\
Pr(H_0 A_1) = Pr_1(H_0 A_1), \ &\text{and} \ Pr(H_0 A_2) = Pr_2(H_0 A_2)
\end{aligned}
$$

*for all atomic events $H_0$, $H_1$, $H_2$, $A_1$, and $A_2$ over the sets of basic events $\{B_0\}$, $\{B_1\}$, $\{B_2\}$, $\mathcal{B}_1$, and $\mathcal{B}_2$, respectively.*

**Proof.** Let the probabilistic interpretation $Pr$ on the atomic events over $\mathcal{B}_1 \cup \mathcal{B}_2 \cup \{B_0\}$ be defined as follows:

$$Pr(H_0 A_1 A_2) = \begin{cases} Pr_0(H_0 H_1 H_2) \cdot \frac{Pr_1(H_0 A_1)}{Pr_1(H_0 H_1)} \cdot \frac{Pr_2(H_0 A_2)}{Pr_2(H_0 H_2)} & \text{if } Pr_1(H_0 H_1) \cdot Pr_2(H_0 H_2) > 0 \\ 0 & \text{if } Pr_1(H_0 H_1) \cdot Pr_2(H_0 H_2) = 0 \end{cases}$$

for all atomic events $H_0$, $A_1$, and $A_2$ over $\{B_0\}$, $\mathcal{B}_1$, and $\mathcal{B}_2$, respectively, with atomic events $H_1$ over $\{B_1\}$ and $H_2$ over $\{B_2\}$ such that $A_1 \Rightarrow H_1$ and $A_2 \Rightarrow H_2$.

Now, we must show that $Pr$ satisfies (12). Let $H_0$, $H_1$, and $H_2$ be atomic events over $\{B_0\}$, $\{B_1\}$, and $\{B_2\}$, respectively. For $Pr_1(H_0 H_1) > 0$ and $Pr_2(H_0 H_2) > 0$, we get:

$$Pr(H_0 H_1 H_2) = \sum_{\substack{A_1 \in \mathcal{A}_{\mathcal{B}_1},\, A_1 \Rightarrow H_1 \\ A_2 \in \mathcal{A}_{\mathcal{B}_2},\, A_2 \Rightarrow H_2}} Pr_0(H_0 H_1 H_2) \cdot \frac{Pr_1(H_0 A_1)}{Pr_1(H_0 H_1)} \cdot \frac{Pr_2(H_0 A_2)}{Pr_2(H_0 H_2)} = Pr_0(H_0 H_1 H_2) \,.$$

For $Pr_1(H_0 H_1) = 0$ or $Pr_2(H_0 H_2) = 0$, we get $Pr(H_0 H_1 H_2) = 0 = Pr_0(H_0 H_1 H_2)$.

Let $H_0$, $H_1$, and $A_1$ be atomic events over $\{B_0\}$, $\{B_1\}$, and $\mathcal{B}_1$, respectively, with $A_1 \Rightarrow H_1$. For $Pr_1(H_0 H_1) > 0$, $Pr_2(H_0 B_2) > 0$, and $Pr_2(H_0 \overline{B}_2) > 0$, it holds:

$$\begin{aligned} Pr(H_0 A_1) \quad &= \sum_{A_2 \in \mathcal{A}_{\mathcal{B}_2},\, A_2 \Rightarrow B_2} Pr_0(H_0 H_1 B_2) \cdot \frac{Pr_1(H_0 A_1)}{Pr_1(H_0 H_1)} \cdot \frac{Pr_2(H_0 A_2)}{Pr_2(H_0 B_2)} \\ &+ \sum_{A_2 \in \mathcal{A}_{\mathcal{B}_2},\, A_2 \Rightarrow \overline{B}_2} Pr_0(H_0 H_1 \overline{B}_2) \cdot \frac{Pr_1(H_0 A_1)}{Pr_1(H_0 H_1)} \cdot \frac{Pr_2(H_0 A_2)}{Pr_2(H_0 \overline{B}_2)} \\ &= Pr_0(H_0 H_1) \cdot \frac{Pr_1(H_0 A_1)}{Pr_1(H_0 H_1)} \qquad = Pr_1(H_0 A_1) \,. \end{aligned}$$

For $Pr_1(H_0 H_1) > 0$, $Pr_2(H_0 B_2) > 0$, and $Pr_2(H_0 \overline{B}_2) = 0$, we get:

$$\begin{aligned} Pr(H_0 A_1) \quad &= \sum_{A_2 \in \mathcal{A}_{\mathcal{B}_2},\, A_2 \Rightarrow B_2} Pr_0(H_0 H_1 B_2) \cdot \frac{Pr_1(H_0 A_1)}{Pr_1(H_0 H_1)} \cdot \frac{Pr_2(H_0 A_2)}{Pr_2(H_0 B_2)} \\ &= Pr_0(H_0 H_1) \cdot \frac{Pr_1(H_0 A_1)}{Pr_1(H_0 H_1)} \qquad = Pr_1(H_0 A_1) \,. \end{aligned}$$

The proof is similar for $Pr_1(H_0 H_1) > 0$, $Pr_2(H_0 B_2) = 0$, and $Pr_2(H_0 \overline{B}_2) > 0$.
For $Pr_1(H_0 H_1) = 0$, we get $Pr(H_0 A_1) = 0 = Pr_1(H_0 A_1)$.

Finally, the proof of $Pr(H_0 A_2) = Pr_2(H_0 A_2)$ for all atomic events $H_0$ over $\{B_0\}$ and $A_2$ over $\mathcal{B}_2$ can be done analogously. $\quad\square$

## Appendix B. Proofs for Section 4.1

In this section, we give the proofs of Theorems 4.1 and 4.2. That is, we show the global soundness and the global completeness of the functions $H_1^\alpha$, $H_2^\alpha$, $H_2^\beta$, and $H_2^\gamma$. The proofs are done by induction on the recursive definition of $H_1^\alpha$, $H_2^\alpha$, $H_2^\beta$, and $H_2^\gamma$.

To prove global soundness, we just have to show the local soundness of the computations in LEAF, CHAINING, and FUSION. To prove global completeness, we construct two models of the conditional constraint tree, one related to the greatest lower bound and another one related to the least upper bound computed in LEAF, CHAINING, and FUSION.

For LEAF, such a model is trivially given. For CHAINING, we combine a model of the arrow, a model of the subtree, and a model connected to the common node to a model of the extended conditional constraint tree. For FUSION, we combine models of the subtrees and a model connected to the common node to a model of the extended conditional constraint tree. More precisely, for CHAINING and FUSION, the models of the subtrees are related to previously computed tightest bounds, while the model connected to the common node is related to the tightest bounds computed in the running CHAINING or FUSION.

We need the following technical preparations. The next lemma helps us to show the global completeness of the functions $H_2^\alpha$, $H_2^\beta$, and $H_2^\gamma$ in CHAINING and FUSION.

**Lemma B.3** a) *For all real numbers $u, v \in (0, 1]$, $x_2 \in [0, 1]$, and $\overline{x}_2, z_2 \in [0, \infty)$ with $x_2, \overline{x}_2 \le z_2$ and $z_2 \le x_2 + \overline{x}_2$, there is some $x \in [z_2 - \overline{x}_2, x_2]$ with:*

$$(13) \qquad \min(1, \tfrac{uz_2}{v}, 1 - u + \tfrac{ux}{v}, u - \tfrac{ux}{v} + \tfrac{uz_2}{v}) \;=\; \min(1, \tfrac{uz_2}{v}, 1 - u + \tfrac{ux_2}{v}, u + \tfrac{u\overline{x}_2}{v}) \,.$$

b) *For $v_2, x_2 \in [0, 1]$ and $\overline{v}_2, \overline{x}_2, w_2, z_2 \in [0, \infty)$ with $v_2 \le w_2$, $\overline{v}_2 \le w_2$, $x_2 \le z_2$, $\overline{x}_2 \le z_2$, $w_2 \le v_2 + \overline{v}_2$, and $z_2 \le x_2 + \overline{x}_2$, there is $v \in [w_2 - \overline{v}_2, v_2]$ and $x \in [z_2 - \overline{x}_2, x_2]$ with:*

$$(14) \qquad \begin{aligned} \min(w_2, z_2, v + z_2 - x, x + w_2 - v) &= \min(w_2, z_2, v_2 + \overline{x}_2, x_2 + \overline{v}_2) \\ \min(v, x) &= \min(v_2, x_2) \,. \end{aligned}$$

**Proof.** The claims can easily be verified (Lukasiewicz, 1996). □

The following lemma helps us to prove the local soundness and the local completeness of the functions $H_1^\alpha$, $H_2^\alpha$, $H_2^\beta$, and $H_2^\gamma$ in CHAINING and FUSION.

**Lemma B.4** a) *Let $u, v \in (0, 1]$, $x \in [0, 1]$, and $\overline{x} \in [0, \infty)$. For all probabilistic interpretations $Pr$ with $Pr(B) > 0$, the conditions $u \cdot Pr(B) = Pr(BC)$, $v \cdot Pr(C) = Pr(BC)$, $x \cdot Pr(C) = Pr(CL(C^\uparrow))$, and $\overline{x} \cdot Pr(C) = Pr(\overline{C}L(C^\uparrow))$ are equivalent to:*

| $\frac{Pr(\overline{B}\,\overline{C}\,\overline{L(C^\uparrow)})}{Pr(B)}$ | $\frac{Pr(\overline{B}\,\overline{C}L(C^\uparrow))}{Pr(B)}$ | $\frac{Pr(\overline{B}\,\overline{C})}{Pr(B)}$ | | $\frac{Pr(\overline{B}\,\overline{CL(C^\uparrow)})}{Pr(B)}$ | $\frac{Pr(\overline{B}\,CL(C^\uparrow))}{Pr(B)}$ | | $\frac{u}{v} - u$ |
|---|---|---|---|---|---|---|---|
| $\frac{Pr(B\overline{C}\,\overline{L(C^\uparrow)})}{Pr(B)}$ | $\frac{Pr(B\,\overline{C}L(C^\uparrow))}{Pr(B)}$ | $1 - u$ | | $\frac{Pr(BC\overline{L(C^\uparrow)})}{Pr(B)}$ | $\frac{Pr(BCL(C^\uparrow))}{Pr(B)}$ | | $u$ |
| $\frac{Pr(\overline{C}\,L(C^\uparrow))}{Pr(B)}$ | $\frac{u\overline{x}}{v}$ | | | $\frac{u}{v} - \frac{ux}{v}$ | $\frac{ux}{v}$ | | |

b) *Let $v, x \in [0, 1]$ and $\overline{v}, \overline{x} \in [0, \infty)$. For all probabilistic interpretations $Pr$ with $Pr(B) > 0$, the conditions $v \cdot Pr(B) = Pr(BL(G))$, $\overline{v} \cdot Pr(B) = Pr(\overline{B}L(G))$, $x \cdot Pr(B) = Pr(BL(H))$, and $\overline{x} \cdot Pr(B) = Pr(\overline{B}L(H))$ are equivalent to:*

| $\frac{Pr(\overline{B}\,\overline{L(G)}\,\overline{L(H)})}{Pr(B)}$ | $\frac{Pr(\overline{B}\,\overline{L(G)}L(H))}{Pr(B)}$ | $\frac{Pr(\overline{B}\,\overline{L(G)})}{Pr(B)}$ | | $\frac{Pr(B\overline{L(G)}\,\overline{L(H)})}{Pr(B)}$ | $\frac{Pr(B\overline{L(G)}L(H))}{Pr(B)}$ | | $1 - v$ |
|---|---|---|---|---|---|---|---|
| $\frac{Pr(\overline{B}L(G)\overline{L(H)})}{Pr(B)}$ | $\frac{Pr(\overline{B}L(G)L(H))}{Pr(B)}$ | $\overline{v}$ | | $\frac{Pr(BL(G)\overline{L(H)})}{Pr(B)}$ | $\frac{Pr(BL(G)L(H))}{Pr(B)}$ | | $v$ |
| $\frac{Pr(\overline{B}\,\overline{L(H)})}{Pr(B)}$ | $\overline{x}$ | | | $1 - x$ | $x$ | | |

**Proof.** The claims can be verified by straightforward arithmetic transformations based on the properties of probabilistic interpretations. □

After these preparations, we are now ready to prove the global soundness and the global completeness of the functions $H_1^\alpha$, $H_2^\alpha$, $H_2^\beta$, and $H_2^\gamma$.

**Proof of Theorem 4.1.** The claims are proved by induction on the recursive definition of $H_1^\alpha$. The case $C = B_1 \ldots B_k$ is tackled by iteratively splitting $C$ into two conjunctive events. Thus, it is reduced to $C = GH$ with conjunctive events $G$ and $H$ that are disjoint in their basic events. For $C = B_1$, we define $u = Pr(C|B)$, $v = Pr(B|C)$, and $x_1 = H_1^\alpha(C, C^\uparrow)$. For $C = B_1 \ldots B_k$, hence $C = GH$, let $v_1 = H_1^\alpha(B, G)$ and $x_1 = H_1^\alpha(B, H)$.

a) All models $Pr \in Mo(B, C)$ with $Pr(B) = 0$ satisfy the indicated condition. In the sequel, let $Pr \in Mo(B, C)$ with $Pr(B) > 0$.

*Basis:* Let $C = B$. Since $C = L(C)$, we get:

$$\alpha_1 \cdot Pr(B) = 1 \cdot Pr(B) = Pr(BC) = Pr(BL(C)).$$

*Induction:* Let $C = B_1$. For all models $Pr_2 \in Mo(C, C^\uparrow)$, we get by the induction hypothesis $x_1 \cdot Pr_2(C) \le Pr_2(CL(C^\uparrow))$. Thus, $Pr$ satisfies the same conditions. Since $L(C^\uparrow) = L(C)$ and by Lemmata A.1 and B.4 a), we then get:

$$\alpha_1 \cdot Pr(B) = \max(0, u - \tfrac{u}{v} + \tfrac{ux_1}{v}) \cdot Pr(B) \le Pr(BL(C^\uparrow)) = Pr(BL(C)).$$

Let $C = GH$. For all $Pr_1 \in Mo(B, G)$ and $Pr_2 \in Mo(B, H)$, we get by the induction hypothesis $v_1 \cdot Pr_1(B) \le Pr_1(BL(G))$ and $x_1 \cdot Pr_2(B) \le Pr_2(BL(H))$. Thus, $Pr$ satisfies the same conditions. Since $L(G)L(H) = L(GH) = L(C)$ and by Lemmata A.1 and B.4 b):

$$\max(0, v_1 + x_1 - 1) \cdot Pr(B) \le Pr(BL(G)L(H)) = Pr(BL(C)).$$

b)

*Basis:* Let $C = B$. A model $Pr \in Mo(B, C)$ such that $Pr(B) > 0$, $1 \cdot Pr(B) = \alpha_1 \cdot Pr(B) = Pr(BL(C))$, and $Pr(\overline{B}L(C)) = 0$ is given by $\overline{B}, B \to 0, 1$.

*Induction:* Let $C = B_1$. Let the model $Pr_1$ of $\{(C|B)[u, u], (B|C)[v, v]\}$ with $Pr_1(B) > 0$ and $Pr_1(C) > 0$ be defined like in the proof of Theorem 3.2.

We now choose an appropriate model $Pr_2 \in Mo(C, C^\uparrow)$. Let us first consider the case $x_1 > 0$, $v = 1$, or not $L(C^\uparrow) \Rightarrow C$. By the induction hypothesis, there exists a model $Pr_2 \in Mo(C, C^\uparrow)$ with $Pr_2(C) > 0$, $x_1 \cdot Pr_2(C) = Pr_2(CL(C^\uparrow))$, and $Pr_2(\overline{C}L(C^\uparrow)) = 0$ iff $L(C^\uparrow) \Rightarrow C$. Let us next assume $x_1 = 0$, $v < 1$, and $L(C^\uparrow) \Rightarrow C$. By Theorem 3.2, there exists a model $Pr_2'' \in Mo(C, C^\uparrow)$ with $Pr_2''(CL(C^\uparrow)) > 0$. By the induction hypothesis, there exists a model $Pr_2' \in Mo(C, C^\uparrow)$ with $Pr_2'(C) > 0$ and $0 \cdot Pr_2'(C) = Pr_2'(CL(C^\uparrow))$. Hence, there exists a model $Pr_2 \in Mo(C, C^\uparrow)$ with $Pr_2(C) > 0$ and

$$\min(1 - v, Pr_2''(CL(C^\uparrow)) / Pr_2''(C)) \cdot Pr_2(C) = Pr_2(CL(C^\uparrow)).$$

By Lemma 3.1, we can choose $Pr_1$ and $Pr_2$ with $Pr_1(C) = Pr_2(C)$ and $Pr_1(\overline{B}\,\overline{C}) \ge Pr_2(\overline{C}L(C^\uparrow))$. By Lemmata A.1 and B.4 a), we can choose the probabilistic interpretation $Pr_0$ over $\{B, C, L(C^{\uparrow})\}$ with $Pr_0(A_1) = Pr_1(A_1)$ and $Pr_0(A_2) = Pr_2(A_2)$ for all atomic events $A_1$ and $A_2$ over $\{B, C\}$ and $\{C, L(C^{\uparrow})\}$, respectively, such that:

$$Pr_0(B\overline{C}L(C^{\uparrow})) = \max(0, Pr_2(\overline{C}L(C^{\uparrow})) - Pr_1(\overline{B}\,\overline{C})) = 0$$
$$Pr_0(BCL(C^{\uparrow})) = \max(0, Pr_2(CL(C^{\uparrow})) - Pr_1(\overline{B}C)) .$$

By Lemma A.2 with $\mathcal{B}_1 = \{B\}$, $\mathcal{B}_2 = \mathcal{B}(C, C^{\uparrow}) \backslash \{C\}$, $B_0 = C$, $B_1 = B$, and $B_2 = L(C^{\uparrow})$, there exists a probabilistic interpretation $Pr$ over $\mathcal{B}(B, C)$ with (12). Hence, it holds $Pr \in Mo(B, C)$ and $Pr(B) > 0$. By Lemma B.4 a), we get:

$$\alpha_1 \cdot Pr(B) = \max(0, u - \tfrac{u}{v} + \tfrac{ux_1}{v}) \cdot Pr(B) = Pr(BL(C^{\uparrow})) = Pr(BL(C)) .$$

Moreover, it is easy to see that $Pr(\overline{B}L(C)) = 0$ iff $L(C) \Rightarrow B$.

Let $C = GH$. By the induction hypothesis, there are models $Pr_1 \in Mo(B, G)$ and $Pr_2 \in Mo(B, H)$ with $Pr_1(B) > 0$, $Pr_2(B) > 0$, $v_1 \cdot Pr_1(B) = Pr_1(BL(G))$, $x_1 \cdot Pr_2(B) = Pr_2(BL(H))$, $Pr_1(\overline{B}L(G)) = 0$ iff $L(G) \Rightarrow B$, and $Pr_2(\overline{B}L(H)) = 0$ iff $L(H) \Rightarrow B$.

By Lemma 3.1, we can choose $Pr_1$ and $Pr_2$ with $Pr_1(B) = Pr_2(B)$ and $Pr_1(\overline{B}\,\overline{L(G)}) \geq Pr_2(\overline{B}L(H))$. By Lemmata A.1 and B.4 b), we can choose the probabilistic interpretation $Pr_0$ over $\{B, L(G), L(H)\}$ with $Pr_0(A_1) = Pr_1(A_1)$ and $Pr_0(A_2) = Pr_2(A_2)$ for all atomic events $A_1$ and $A_2$ over $\{B, L(G)\}$ and $\{B, L(H)\}$, respectively, such that:

$$Pr_0(\overline{B}L(G)L(H)) = \min(Pr_2(\overline{B}L(H)), Pr_1(\overline{B}L(G))$$
$$Pr_0(BL(G)L(H)) = \max(0, Pr_2(BL(H)) - Pr_1(B\overline{L(G)})) .$$

By Lemma A.2 with $\mathcal{B}_1 = \mathcal{B}(B, G) \backslash \{B\}$, $\mathcal{B}_2 = \mathcal{B}(B, H) \backslash \{B\}$, $B_0 = B$, $B_1 = L(G)$, and $B_2 = L(H)$, there exists a probabilistic interpretation $Pr$ over $\mathcal{B}(B, C)$ with (12). Hence, it holds $Pr \in Mo(B, C)$ and $Pr(B) > 0$. By Lemma B.4 b), we get:

$$\max(0, v_1 + x_1 - 1) \cdot Pr(B) = Pr(BL(G)L(H)) = Pr(BL(C)) .$$

Moreover, it is easy to see that $Pr(\overline{B}L(C)) = 0$ iff $L(C) \Rightarrow B$. $\square$

**Proof of Theorem 4.2.** The claims are proved by induction on the recursive definition of $H_2^{\alpha}$, $H_2^{\beta}$, and $H_2^{\gamma}$. Again, the case $C = B_1 \ldots B_k$ is tackled by iteratively splitting $C$ into two conjunctive events. Thus, it is reduced to $C = GH$ with conjunctive events $G$ and $H$ that are disjoint in their basic events. For $C = B_1$ let $u = Pr(C|B)$, $v = Pr(B|C)$, and

$$x_2 = H_2^{\alpha}(C, C^{\uparrow}), \quad \overline{x}_2 = H_2^{\beta}(C, C^{\uparrow}), \quad z_2 = H_2^{\gamma}(C, C^{\uparrow}) .$$

For $C = B_1 \ldots B_k$, hence $C = GH$, we define:

$$v_2 = H_2^{\alpha}(B, G), \quad \overline{v}_2 = H_2^{\beta}(B, G), \quad w_2 = H_2^{\gamma}(B, G)$$
$$x_2 = H_2^{\alpha}(B, H), \quad \overline{x}_2 = H_2^{\beta}(B, H), \quad z_2 = H_2^{\gamma}(B, H) .$$

a) For $Pr \in Mo(B, C)$ with $Pr(B) = 0$, we get $Pr(N) = 0$ for all $N \in \mathcal{B}(B, C)$. Thus, $Pr$ satisfies the indicated conditions. Next, let $Pr \in Mo(B, C)$ with $Pr(B) > 0$.

*Basis:* Let $C = B$. Since $L(C) = C$, we get:

$$Pr(BL(C)) = Pr(BC) = 1 \cdot Pr(B) = \alpha_2 \cdot Pr(B)$$
$$Pr(\overline{B}L(C)) = Pr(\overline{B}C) = 0 \cdot Pr(B) = \beta_2 \cdot Pr(B)$$
$$Pr(L(C)) = Pr(C) = 1 \cdot Pr(B) = \gamma_2 \cdot Pr(B) .$$

*Induction:* Let $C = B_1$. For all models $Pr_2 \in Mo(C, C^\uparrow)$, we get by the induction hypothesis $Pr_2(CL(C^\uparrow)) \leq x_2 \cdot Pr_2(C)$, $Pr_2(\overline{C}L(C^\uparrow)) \leq \overline{x}_2 \cdot Pr_2(C)$, and $Pr_2(L(C^\uparrow)) \leq z_2 \cdot Pr_2(C)$. Hence, $Pr$ satisfies the same conditions. Since $L(C) = L(C^\uparrow)$ and by Lemmata A.1 and B.4 a), we then get:

$$Pr(BL(C)) = Pr(BL(C^\uparrow)) \leq \min(1, \tfrac{uz_2}{v}, 1 - u + \tfrac{ux_2}{v}, u + \tfrac{u\overline{x}_2}{v}) \cdot Pr(B) = \alpha_2 \cdot Pr(B)$$
$$Pr(\overline{B}L(C)) = Pr(\overline{B}L(C^\uparrow)) \leq \min(\tfrac{u\overline{x}_2}{v} + \tfrac{u}{v} - u, \tfrac{uz_2}{v}) \cdot Pr(B) = \beta_2 \cdot Pr(B)$$
$$Pr(L(C)) = Pr(L(C^\uparrow)) \leq \tfrac{uz_2}{v} \cdot Pr(B) = \gamma_2 \cdot Pr(B) .$$

Let $C = GH$. For all models $Pr_1 \in Mo(B, G)$ and all models $Pr_2 \in Mo(B, H)$, we get by the induction hypothesis:

$$Pr_1(BL(G)) \leq v_2 \cdot Pr_1(B), \quad Pr_2(BL(H)) \leq x_2 \cdot Pr_2(B)$$
$$Pr_1(\overline{B}L(G)) \leq \overline{v}_2 \cdot Pr_1(B), \quad Pr_2(\overline{B}L(H)) \leq \overline{x}_2 \cdot Pr_2(B)$$
$$Pr_1(L(G)) \leq w_2 \cdot Pr_1(B), \quad Pr_2(L(H)) \leq z_2 \cdot Pr_2(B) .$$

Thus, $Pr$ satisfies the same conditions. Since $L(C) = L(GH) = L(G)L(H)$ and by Lemmata A.1 and B.4 b), we get:

$$Pr(BL(C)) = Pr(BL(G)L(H)) \leq \min(v_2, x_2) \cdot Pr(B)$$
$$Pr(\overline{B}L(C)) = Pr(\overline{B}L(G)L(H)) \leq \min(\overline{v}_2, \overline{x}_2) \cdot Pr(B)$$
$$Pr(L(C)) = Pr(L(G)L(H)) \leq \min(w_2, z_2, v_2 + \overline{x}_2, x_2 + \overline{v}_2) \cdot Pr(B) .$$

b) and c)

*Basis:* Let $C = B$. A model $Pr \in Mo(B, C)$ with $Pr(B) > 0$ satisfying $Pr(BL(C)) = 1 \cdot Pr(B) = \alpha_2 \cdot Pr(B)$, $Pr(\overline{B}L(C)) = 0 \cdot Pr(B) = \beta_2 \cdot Pr(B)$, and $Pr(L(C)) = 1 \cdot Pr(B) = \gamma_2 \cdot Pr(B)$ is given by $\overline{B}, B \mapsto 0, 1$.

*Induction:* Let $C = B_1$. Let the model $Pr_1$ of $\{(C|B)[u, u], (B|C)[v, v]\}$ with $Pr_1(B) > 0$ and $Pr_1(C) > 0$ be defined like in the proof of Theorem 3.2.

For the proof of c), by the induction hypothesis, there is some $Pr_2 \in Mo(C, C^\uparrow)$ with $Pr_2(C) > 0$, $Pr_2(\overline{C}L(C^\uparrow)) = \overline{x}_2 \cdot Pr_2(C)$, and $Pr_2(L(C^\uparrow)) = z_2 \cdot Pr_2(C)$.

By Lemma 3.1, we can choose $Pr_1$ and $Pr_2$ with $Pr_1(C) = Pr_2(C)$ and $Pr_1(\overline{B}\,\overline{C}) \geq Pr_2(\overline{C}L(C^\uparrow))$. By Lemma A.1, we can choose the probabilistic interpretation $Pr_0$ over $\{B, C, L(C^\uparrow)\}$ with $Pr_0(A_1) = Pr_1(A_1)$ and $Pr_0(A_2) = Pr_2(A_2)$ for all atomic events $A_1$ and $A_2$ over $\{B, C\}$ and $\{C, L(C^\uparrow)\}$, respectively, such that:

$$Pr_0(\overline{B}\,\overline{C}L(C^\uparrow)) = \min(Pr_1(\overline{B}\,\overline{C}), Pr_2(\overline{C}L(C^\uparrow))) = Pr_2(\overline{C}L(C^\uparrow))$$
$$Pr_0(\overline{B}CL(C^\uparrow)) = \min(Pr_1(\overline{B}C), Pr_2(CL(C^\uparrow))) .$$

By Lemma A.2 with $\mathcal{B}_1 = \{B\}$, $\mathcal{B}_2 = \mathcal{B}(C, C^\uparrow) \setminus \{C\}$, $B_0 = C$, $B_1 = B$, and $B_2 = L(C^\uparrow)$, there is a probabilistic interpretation $Pr$ over $\mathcal{B}(B, C)$ with (12). Hence, it holds $Pr \in Mo(B, C)$ and $Pr(B) > 0$. By Lemma B.4 a), we get:

$$Pr(\overline{B}L(C)) \;=\; Pr(\overline{B}L(C^\uparrow)) \;=\; \min(\tfrac{u\overline{x}_2}{v} + \tfrac{u}{v} - u, \tfrac{uz_2}{v}) \cdot Pr(B) \;=\; \beta_2 \cdot Pr(B)$$
$$Pr(L(C)) \;=\; Pr(L(C^\uparrow)) \;=\; \tfrac{uz_2}{v} \cdot Pr(B) \;=\; \gamma_2 \cdot Pr(B) \;.$$

For the proof of b), by the induction hypothesis, there are models $Pr_{1,2}$, $Pr_{2,2} \in Mo(C, C^\uparrow)$ with $Pr_{1,2}(C) > 0$, $Pr_{2,2}(C) > 0$, and

$$(15) \qquad \begin{aligned} Pr_{1,2}(CL(C^\uparrow)) &= x_2 \cdot Pr_{1,2}(C), \quad Pr_{1,2}(L(C^\uparrow)) = z_2 \cdot Pr_{1,2}(C) \\ Pr_{2,2}(\overline{C}L(C^\uparrow)) &= \overline{x}_2 \cdot Pr_{2,2}(C), \quad Pr_{2,2}(L(C^\uparrow)) = z_2 \cdot Pr_{2,2}(C) \;. \end{aligned}$$

These conditions already entail $x_2 \leq z_2$ and $\overline{x}_2 \leq z_2$. With the results from a), we additionally get $z_2 \leq x_2 + \overline{x}_2$. By Lemma B.3 a), there is $x \in [z_2 - \overline{x}_2, x_2]$ with (13). By (15), there is $Pr_2 \in Mo(C, C^\uparrow)$ with $Pr_2(C) > 0$ and

$$Pr_2(CL(C^\uparrow)) = x \cdot Pr_2(C), \quad Pr_2(L(C^\uparrow)) = z_2 \cdot Pr_2(C) \;.$$

By Lemma 3.1, we can choose $Pr_1$ and $Pr_2$ with $Pr_1(C) = Pr_2(C)$. By Lemma A.1, we can choose the probabilistic interpretation $Pr_0$ over $\{B, C, L(C^\uparrow)\}$ with $Pr_0(A_1) = Pr_1(A_1)$ and $Pr_0(A_2) = Pr_2(A_2)$ for all atomic events $A_1$ and $A_2$ over $\{B, C\}$ and $\{C, L(C^\uparrow)\}$, respectively, such that:

$$Pr_0(B\overline{C}L(C^\uparrow)) \;=\; \min(Pr_1(B\overline{C}), Pr_2(\overline{C}L(C^\uparrow)))$$
$$Pr_0(BCL(C^\uparrow)) \;=\; \min(Pr_1(BC), Pr_2(CL(C^\uparrow))) \;.$$

By Lemma A.2 with $\mathcal{B}_1 = \{B\}$, $\mathcal{B}_2 = \mathcal{B}(C, C^\uparrow) \setminus \{C\}$, $B_0 = C$, $B_1 = B$, and $B_2 = L(C^\uparrow)$, there is a probabilistic interpretation $Pr$ over $\mathcal{B}(B, C)$ with (12). Hence, it holds $Pr \in Mo(B, C)$ and $Pr(B) > 0$. By Lemma B.4 a), we get:

$$Pr(BL(C)) \;=\; Pr(BL(C^\uparrow)) \;=\; \min(1, \tfrac{uz_2}{v}, 1 - u + \tfrac{ux_2}{v}, u + \tfrac{u\overline{x}_2}{v}) \cdot Pr(B) \;=\; \alpha_2 \cdot Pr(B)$$
$$Pr(L(C)) \;=\; Pr(L(C^\uparrow)) \;=\; \tfrac{uz_2}{v} \cdot Pr(B) \;=\; \gamma_2 \cdot Pr(B) \;.$$

Let $C = GH$. We just show b), the claim in c) can be proved analogously. By the induction hypothesis, there are models $Pr_{1,1}$, $Pr_{2,1} \in Mo(B, G)$ and $Pr_{1,2}$, $Pr_{2,2} \in Mo(B, H)$ with $Pr_{1,1}(B) > 0$, $Pr_{2,1}(B) > 0$, $Pr_{1,2}(B) > 0$, $Pr_{2,2}(B) > 0$, and

$$(16) \qquad \begin{aligned} Pr_{1,1}(BL(G)) &= v_2 \cdot Pr_{1,1}(B), \quad Pr_{1,1}(L(G)) = w_2 \cdot Pr_{1,1}(B) \\ Pr_{2,1}(\overline{B}L(G)) &= \overline{v}_2 \cdot Pr_{2,1}(B), \quad Pr_{2,1}(L(G)) = w_2 \cdot Pr_{2,1}(B) \\ Pr_{1,2}(BL(H)) &= x_2 \cdot Pr_{1,2}(B), \quad Pr_{1,2}(L(H)) = z_2 \cdot Pr_{1,2}(B) \\ Pr_{2,2}(\overline{B}L(H)) &= \overline{x}_2 \cdot Pr_{2,2}(B), \quad Pr_{2,2}(L(H)) = z_2 \cdot Pr_{2,2}(B) \;. \end{aligned}$$

These conditions already entail $v_2 \leq w_2$, $\overline{v}_2 \leq w_2$, $x_2 \leq z_2$, and $\overline{x}_2 \leq z_2$. With the results from a), we additionally get $w_2 \leq v_2 + \overline{v}_2$ and $z_2 \leq x_2 + \overline{x}_2$. By Lemma B.3 b), there is $v \in [w_2 - \overline{v}_2, v_2]$ and $x \in [z_2 - \overline{x}_2, x_2]$ with (14). By (16), there is $Pr_1 \in Mo(B, G)$ and $Pr_2 \in Mo(B, H)$ with $Pr_1(B) > 0$, $Pr_2(B) > 0$, and

$$Pr_1(BL(G)) = v \cdot Pr_1(B), \quad Pr_1(L(G)) = w_2 \cdot Pr_1(B)$$
$$Pr_2(BL(H)) = x \cdot Pr_2(B), \quad Pr_2(L(H)) = z_2 \cdot Pr_2(B) \ .$$

By Lemma 3.1, we can choose $Pr_1$ and $Pr_2$ with $Pr_1(B) = Pr_2(B)$. By Lemma A.1, we can choose the probabilistic interpretation $Pr_0$ over $\{B, L(G), L(H)\}$ with $Pr_0(A_1) = Pr_1(A_1)$ and $Pr_0(A_2) = Pr_2(A_2)$ for all atomic events $A_1$ and $A_2$ over $\{B, L(G)\}$ and $\{B, L(H)\}$, respectively, such that:

$$Pr_0(\overline{B}L(G)L(H)) = \min(Pr_1(\overline{B}L(G)), Pr_2(\overline{B}L(H)))$$
$$Pr_0(BL(G)L(H)) = \min(Pr_1(BL(G)), Pr_2(BL(H))) \ .$$

By Lemma A.2 with $\mathcal{B}_1 = \mathcal{B}(B, G) \setminus \{B\}$, $\mathcal{B}_2 = \mathcal{B}(B, H) \setminus \{B\}$, $B_0 = B$, $B_1 = L(G)$, and $B_2 = L(H)$, there is a probabilistic interpretation $Pr$ over $\mathcal{B}(B, C)$ with (12). Hence, it holds $Pr \in Mo(B, C)$ and $Pr(B) > 0$. By Lemma B.4 b), we get:

$$Pr(BL(C)) = Pr(BL(G)L(H)) = \min(v_2, x_2) \cdot Pr(B)$$
$$Pr(L(C)) = Pr(L(G)L(H)) = \min(w_2, z_2, v_2 + \overline{x}_2, x_2 + \overline{v}_2) \cdot Pr(B) \ . \quad \square$$

Finally, note that computing least upper bounds is more difficult than computing greatest lower bounds, since for each edge $B \to C$, by Lemmata 3.1 and B.4 a), the greatest lower bound of $Pr(B\overline{C}L(C^{\uparrow}))/Pr(B)$ subject to $Pr \in Mo(B, C)$ and $Pr(B) > 0$ is always 0, but the least upper bound of $Pr(B\overline{C}L(C^{\uparrow}))/Pr(B)$ subject to $Pr \in Mo(B, C)$ and $Pr(B) > 0$ is generally not 1.

## Appendix C. Proofs for Section 4.2

In this section, we give the proofs of Theorems 4.7 and 4.8.

We need some technical preparations as follows. The next lemma helps us to show the local soundness of the function $H_1^{\delta}$ in FUSION.

**Lemma C.5** *For all real numbers $u_1, u, v_1, v, x_1, x, y_1, y \in (0, 1]$ with $u_1 \leq u$, $v_1 \leq v$, $x_1 \leq x$, $y_1 \leq y$, and $u_1 + x_1 > 1$, it holds:*

$$\min(u/v - u, x/y - x) / (u + x - 1) \leq \min(u_1/v_1 - u_1, x_1/y_1 - x_1) / (u_1 + x_1 - 1) \ .$$

**Proof.** The claim can easily be verified (Lukasiewicz, 1996). $\square$

The following lemma helps us to show the local soundness and the local completeness of the function $H_1^{\delta}$ in CHAINING and FUSION.

**Lemma C.6** *a) Let $u$, $v$, $x$, and $y$ be real numbers from $(0, 1]$. For all probabilistic interpretations $Pr$ with $Pr(L(C^{\uparrow})) > 0$, the conditions $u \cdot Pr(B) = Pr(BC)$, $v \cdot Pr(C) = Pr(BC)$, $x \cdot Pr(C) = Pr(CL(C^{\uparrow}))$, and $y \cdot Pr(L(C^{\uparrow})) = Pr(CL(C^{\uparrow}))$ are equivalent to:*

| | | | | | |
|---|---|---|---|---|---|
| $\dfrac{Pr(\overline{B}\,\overline{C}\,L(C^\uparrow))}{Pr(L(C^\uparrow))}$ | $\dfrac{Pr(\overline{B}\,\overline{C}L(C^\uparrow))}{Pr(L(C^\uparrow))}$ | $\dfrac{Pr(\overline{B}\,\overline{C})}{Pr(L(C^\uparrow))}$ | $\dfrac{Pr(\overline{B}\,C\overline{L(C^\uparrow)})}{Pr(L(C^\uparrow))}$ | $\dfrac{Pr(\overline{B}\,CL(C^\uparrow))}{Pr(L(C^\uparrow))}$ | $\dfrac{y}{x}-\dfrac{yv}{x}$ |
| $\dfrac{Pr(B\overline{C}\,\overline{L(C^\uparrow)})}{Pr(L(C^\uparrow))}$ | $\dfrac{Pr(B\overline{C}L(C^\uparrow))}{Pr(L(C^\uparrow))}$ | $\dfrac{yv}{xu}-\dfrac{yv}{x}$ | $\dfrac{Pr(BC\overline{L(C^\uparrow)})}{Pr(L(C^\uparrow))}$ | $\dfrac{Pr(BCL(C^\uparrow))}{Pr(L(C^\uparrow))}$ | $\dfrac{yv}{x}$ |
| $\dfrac{Pr(\overline{C}\,L(C^\uparrow))}{Pr(L(C^\uparrow))}$ | $1-y$ | | $\dfrac{y}{x}-y$ | $y$ | |

b) *Let $u$, $v$, $x$, and $y$ be real numbers from $(0,1]$. For all probabilistic interpretations $Pr$ with $Pr(B) > 0$, the conditions $u \cdot Pr(B) = Pr(BL(G))$, $v \cdot Pr(L(G)) = Pr(BL(G))$, $x \cdot Pr(B) = Pr(BL(H))$, and $y \cdot Pr(L(H)) = Pr(BL(H))$ are equivalent to:*

| | | | | | |
|---|---|---|---|---|---|
| $\dfrac{Pr(\overline{B}\,L(G)\,L(H))}{Pr(B)}$ | $\dfrac{Pr(\overline{B}\,L(G)L(H))}{Pr(B)}$ | $\dfrac{Pr(\overline{B}\,L(G))}{Pr(B)}$ | $\dfrac{Pr(B\overline{L(G)}\,L(H))}{Pr(B)}$ | $\dfrac{Pr(B\overline{L(G)}L(H))}{Pr(B)}$ | $1-u$ |
| $\dfrac{Pr(\overline{B}L(G)\overline{L(H)})}{Pr(B)}$ | $\dfrac{Pr(\overline{B}L(G)L(H))}{Pr(B)}$ | $\dfrac{u}{v}-u$ | $\dfrac{Pr(BL(G)\overline{L(H)})}{Pr(B)}$ | $\dfrac{Pr(BL(G)L(H))}{Pr(B)}$ | $u$ |
| $\dfrac{Pr(\overline{B}\,L(H))}{Pr(B)}$ | $\dfrac{x}{y}-x$ | | $1-x$ | $x$ | |

**Proof.** The claims can be verified by straightforward arithmetic transformations based on the properties of probabilistic interpretations. $\square$

We are now ready to prove Theorems 4.7 and 4.8.

**Proof of Theorem 4.7.** The claims are proved by induction on the recursive definition of $H_1^\delta$. The proof for $C = B_1 B_2 \ldots B_k$ with $k > 1$ is done for $k = 2$. It can easily be generalized to $k \geq 2$. For $C = B_1$, we define $u_1 = Pr_1(C|B)$, $v_1 = Pr_1(B|C)$, $x_1 = H_1^\alpha(C, C^\uparrow)$, and $y_1 = H_1^\delta(C, C^\uparrow)$. Note that $\alpha_1 > 0$ entails $x_1, y_1 > 0$ and $v_1 + x_1 > 1$. For $C = B_1 B_2$, we define $G = B_1$, $H = B_2$, $u_1 = H_1^\alpha(B, G)$, $v_1 = H_1^\delta(B, G)$, $x_1 = H_1^\alpha(B, H)$, and $y_1 = H_1^\delta(B, H)$. Note that $\alpha_1 > 0$ entails $u_1, v_1, x_1, y_1 > 0$ and $u_1 + x_1 > 1$.

a) All models $Pr \in Mo(B, C)$ with $Pr(L(C)) = 0$ satisfy the indicated condition. In the sequel, let $Pr \in Mo(B, C)$ with $Pr(L(C)) > 0$ and thus also $Pr(B) > 0$.

*Basis:* Let $C = B$. Since $C = L(C)$, we get:

$$\delta_1 \cdot Pr(L(C)) \;=\; 1 \cdot Pr(L(C)) \;=\; Pr(CL(C)) \;=\; Pr(BL(C)) \,.$$

*Induction:* Let $C = B_1$. For all models $Pr_2 \in Mo(C, C^\uparrow)$, we get $x_1 \cdot Pr_2(C) \leq Pr_2(CL(C^\uparrow))$ by Theorem 4.3 a), and $y_1 \cdot Pr_2(L(C^\uparrow)) \leq Pr_2(CL(C^\uparrow))$ by the induction hypothesis. Thus, $Pr$ satisfies the same conditions. Since $L(C^\uparrow) = L(C)$ and by Lemmata A.1 and C.6 a):

$$\delta_1 \;=\; y_1 - \frac{y_1}{x_1} + \frac{y_1 v_1}{x_1} \;\leq\; Pr(CL(C^\uparrow)) \,/\, Pr(L(C^\uparrow)) \;=\; Pr(CL(C)) \,/\, Pr(L(C)) \,.$$

Let $C = GH$. For all models $Pr_1 \in Mo(B, G)$ and $Pr_2 \in Mo(B, H)$, we get by Theorem 4.3 a) and by the induction hypothesis, respectively:

$$u_1 \cdot Pr_1(B) \;\leq\; Pr_1(BL(G)), \qquad x_1 \cdot Pr_2(B) \;\leq\; Pr_2(BL(H))$$
$$v_1 \cdot Pr_1(L(G)) \;\leq\; Pr_1(BL(G)), \quad y_1 \cdot Pr_2(L(H)) \;\leq\; Pr_2(BL(H)) \,.$$

Hence, $Pr$ satisfies the same conditions. Since $L(G)L(H) = L(GH) = L(C)$ and by Lemmata A.1, C.5, and C.6 b), we then get:

$$\delta_1 = 1 / (1 + \tfrac{\min(u_1/v_1 - u_1, x_1/y_1 - x_1)}{u_1 + x_1 - 1}) \leq 1 / (1 + \tfrac{Pr(\overline{B}L(G)L(H))/Pr(B)}{Pr(BL(G)L(H))/Pr(B)}) = \tfrac{Pr(BL(C))}{Pr(L(C))} .$$

b)

*Basis:* Let $C = B$. A model $Pr \in Mo(B, C)$ such that $Pr(B) > 0$, $Pr(L(C)) > 0$, $1 \cdot Pr(L(C)) = Pr(BL(C))$, and $1 \cdot Pr(B) = Pr(BL(C))$ is given by $\overline{B}, B \mapsto 0, 1$.

*Induction:* Let $C = B_1$. Let the model $Pr_1$ of $\{(C|B)[u_1, u_1], (B|C)[v_1, v_1]\}$ with $Pr_1(B) > 0$ and $Pr_1(C) > 0$ be defined like in the proof of Theorem 3.2.

By the induction hypothesis, there is $Pr_2 \in Mo(C, C^\uparrow)$ with $Pr_2(C) > 0$, $Pr_2(L(C^\uparrow)) > 0$, $y_1 \cdot Pr_2(L(C^\uparrow)) = Pr_2(CL(C^\uparrow))$, and $x_1 \cdot Pr_2(C) = Pr_2(CL(C^\uparrow))$. By Lemma 3.1, we can choose $Pr_1$ and $Pr_2$ such that $Pr_1(C) = Pr_2(C)$ and $Pr_1(\overline{B}\,\overline{C}) \geq Pr_2(\overline{C}L(C^\uparrow))$. By Lemmata A.1 and C.6 a), we can choose the probabilistic interpretation $Pr_0$ over $\{B, C, L(C^\uparrow)\}$ with $Pr_0(A_1) = Pr_1(A_1)$ and $Pr_0(A_2) = Pr_2(A_2)$ for all atomic events $A_1$ and $A_2$ over $\{B, C\}$ and $\{C, L(C^\uparrow)\}$, respectively, such that:

$$Pr_0(B\overline{C}L(C^\uparrow)) = \max(0, Pr_2(\overline{C}L(C^\uparrow)) - Pr_1(\overline{B}\,\overline{C})) = 0$$
$$Pr_0(BCL(C^\uparrow)) = \max(0, Pr_2(CL(C^\uparrow)) - Pr_1(\overline{B}C)) .$$

By Lemma A.2 with $\mathcal{B}_1 = \{B\}$, $\mathcal{B}_2 = \mathcal{B}(C, C^\uparrow) \backslash \{C\}$, $B_0 = C$, $B_1 = B$, and $B_2 = L(C^\uparrow)$, there exists a probabilistic interpretation $Pr$ over $\mathcal{B}(B, C)$ with (12). Hence, it holds $Pr \in Mo(B, C)$, $Pr(B) > 0$, and $Pr(L(C)) > 0$. Moreover, by Lemma C.6 a), we get:

$$\delta_1 = y_1 - \tfrac{y_1}{x_1} + \tfrac{y_1 v_1}{x_1} = Pr(CL(C^\uparrow)) / Pr(L(C^\uparrow)) = Pr(CL(C)) / Pr(L(C))$$
$$\alpha_1 = u_1 - \tfrac{u_1}{v_1} + \tfrac{u_1 x_1}{v_1} = Pr(CL(C^\uparrow)) / Pr(C) = Pr(CL(C)) / Pr(C) .$$

Let $C = GH$. By the induction hypothesis, there are models $Pr_1 \in Mo(B, G)$ and $Pr_2 \in Mo(B, H)$ with $Pr_1(B) > 0$, $Pr_2(B) > 0$, $Pr_1(L(G)) > 0$, $Pr_2(L(H)) > 0$, and

$$u_1 \cdot Pr_1(B) = Pr_1(BL(G)), \qquad x_1 \cdot Pr_2(B) = Pr_2(BL(H))$$
$$v_1 \cdot Pr_1(L(G)) = Pr_1(BL(G)), \quad y_1 \cdot Pr_2(L(H)) = Pr_2(BL(H)) .$$

By Lemma 3.1, we can choose $Pr_1$ and $Pr_2$ with $Pr_1(B) = Pr_2(B)$ and $Pr_1(\overline{B}\,\overline{L(G)}) \geq Pr_2(\overline{B}L(H)$. By Lemmata A.1 and C.6 b), we can choose the probabilistic interpretation $Pr_0$ over $\{B, L(G), L(H)\}$ with $Pr_0(A_1) = Pr_1(A_1)$ and $Pr_0(A_2) = Pr_2(A_2)$ for all atomic events $A_1$ and $A_2$ over $\{B, L(G)\}$ and $\{B, L(H)\}$, respectively, such that:

$$Pr_0(\overline{B}L(G)L(H)) = \min(Pr_2(\overline{B}L(H)), Pr_1(\overline{B}L(G)))$$
$$Pr_0(BL(G)L(H)) = \max(0, Pr_2(BL(H)) - Pr_1(B\overline{L(G)})) .$$

By Lemma A.2 with $\mathcal{B}_1 = \mathcal{B}(B, G) \backslash \{B\}$, $\mathcal{B}_2 = \mathcal{B}(B, H) \backslash \{B\}$, $B_0 = B$, $B_1 = L(G)$, and $B_2 = L(H)$, there exists a probabilistic interpretation $Pr$ over $\mathcal{B}(B, C)$ with (12). Hence, it holds $Pr \in Mo(B, C)$ and $Pr(B) > 0$. By Lemma C.6 b), we get $Pr(L(C)) > 0$ and

$$\delta_1 = 1 / (1 + \frac{\min(u_1/v_1 - u_1, x_1/y_1 - x_1)}{u_1 + x_1 - 1}) = 1 / (1 + \frac{Pr(\overline{B}L(G)L(H))/Pr(B)}{Pr(BL(G)L(H))/Pr(B)}) = \frac{Pr(BL(C))}{Pr(L(C))}$$

$$\alpha_1 = u_1 + x_1 - 1 = \frac{Pr(BL(G)L(H))}{Pr(B)} = \frac{Pr(BL(C))}{Pr(B)} .$$

c) Let $C = GH$. By Theorem 4.3 b), there exist $Pr_1 \in Mo(B, G)$ and $Pr_2 \in Mo(B, H)$ with $Pr_1(B) > 0$, $Pr_2(B) > 0$, $u_1 \cdot Pr_1(B) = Pr_1(BL(G))$, and $x_1 \cdot Pr_2(B) = Pr_2(BL(H))$. By Lemma 3.1, we can choose $Pr_1$ and $Pr_2$ such that $Pr_1(B) = Pr_2(B)$ and $Pr_1(\overline{B}\,\overline{L(G)}) \geq Pr_2(\overline{B}L(H))$. By Lemmata A.1 and C.6 b), we can choose the probabilistic interpretation $Pr_0$ over $\{B, L(G), L(H)\}$ with $Pr_0(A_1) = Pr_1(A_1)$ and $Pr_0(A_2) = Pr_2(A_2)$ for all atomic events $A_1$ and $A_2$ over $\{B, L(G)\}$ and $\{B, L(H)\}$, respectively, such that:

$$Pr_0(\overline{B}L(G)L(H)) = \max(0, Pr_2(\overline{B}L(H) - Pr_1(\overline{B}\,\overline{L(G)}))) = 0$$
$$Pr_0(BL(G)L(H)) = \max(0, Pr_2(BL(H)) - Pr_1(B\overline{L(G)})) .$$

By Lemma A.2 with $\mathcal{B}_1 = \mathcal{B}(B, G) \setminus \{B\}$, $\mathcal{B}_2 = \mathcal{B}(B, H) \setminus \{B\}$, $B_0 = B$, $B_1 = L(G)$, and $B_2 = L(H)$, there exists a probabilistic interpretation $Pr$ over $\mathcal{B}(B, C)$ with (12). Hence, it holds $Pr \in Mo(B, C)$ and $Pr(B) > 0$. By Lemma C.6 b), we get $Pr(L(C)) > 0$ and

$$1 = Pr(BL(G)L(H)) / Pr(L(G)L(H)) = Pr(BL(C)) / Pr(L(C))$$
$$\alpha_1 = Pr(BL(G)L(H)) / Pr(B)) = Pr(BL(C)) / Pr(B) .$$

d) Let $C = GH$. By Theorem 3.2, there is a model $Pr''' \in Mo(B, C)$ with $Pr'''(BL(C)) > 0$. By Theorem 4.3 b), there is a model $Pr'' \in Mo(B, C)$ with $Pr''(B) > 0$ and $0 \cdot Pr''(B) = \alpha_1 \cdot Pr''(B) = Pr''(BL(C))$. Hence, there is a model $Pr' \in Mo(B, C)$ with $Pr'(B) > 0$ and

$$\min(\varepsilon, Pr'''(BL(C)) / Pr'''(B)) \cdot Pr'(B) = Pr'(BL(C)) .$$

Let the models $Pr_1 \in Mo(B, G)$ and $Pr_2 \in Mo(B, H)$ be defined by $Pr_1(A_1) = Pr'(A_1)$ and $Pr_2(A_2) = Pr'(A_2)$ for all atomic events $A_1$ and $A_2$ over $\mathcal{B}(B, G)$ and $\mathcal{B}(B, H)$, respectively. By Lemma 3.1, we can choose $Pr_1$ and $Pr_2$ such that $Pr_1(B) = Pr_2(B)$ and $Pr_1(\overline{B}\,\overline{L(G)}) \geq Pr_2(\overline{B}L(H))$. By Lemmata A.1 and C.6 b), we can choose the probabilistic interpretation $Pr_0$ over $\{B, L(G), L(H)\}$ with $Pr_0(A_1) = Pr_1(A_1)$ and $Pr_0(A_2) = Pr_2(A_2)$ for all atomic events $A_1$ and $A_2$ over $\{B, L(G)\}$ and $\{B, L(H)\}$, respectively, such that:

$$Pr_0(\overline{B}L(G)L(H)) = \max(0, Pr_2(\overline{B}L(H) - Pr_1(\overline{B}\,\overline{L(G)}))) = 0$$
$$Pr_0(BL(G)L(H)) = \min(\varepsilon, Pr'''(BL(C)) / Pr'''(B)) \cdot Pr_0(B) .$$

By Lemma A.2 with $\mathcal{B}_1 = \mathcal{B}(B, G) \setminus \{B\}$, $\mathcal{B}_2 = \mathcal{B}(B, H) \setminus \{B\}$, $B_0 = B$, $B_1 = L(G)$, and $B_2 = L(H)$, there is a probabilistic interpretation $Pr$ over $\mathcal{B}(B, C)$ with (12). Hence, it holds $Pr \in Mo(B, C)$, $Pr(B) > 0$, $Pr(L(C)) > 0$, $Pr(\overline{B}L(C)) = 0$, and $\varepsilon \cdot Pr(B) \geq Pr(BL(C))$. □

**Proof of Theorem 4.8.** For $u_1 > 0$, the claim is immediate by Theorem 4.7 a) to c).

Let $u_1 = 0$ and $E \Rightarrow F$. It holds $1 \cdot Pr(E) = Pr(EF)$ for all models $Pr$ of $KB$. Moreover, by Theorem 3.2, there exists a model $Pr$ of $KB$ with $Pr(E) > 0$.

Let $u_1 = 0$ and not $E \Rightarrow F$. By Theorem 4.3 b), there exists a model $Pr$ of $KB$ with $Pr(E) > 0$ and $Pr(EF) = 0$. By Theorem 4.7 d), there exists a model $Pr$ of $KB$ with $Pr(E) > 0$ and $1 \cdot Pr(E) = Pr(EF)$. □

## Appendix D. Proofs for Section 4.3

In this section, we give the proof of Theorem 4.9.

The next lemma will help us to show the global tightness of the computed lower bound in the case (3) of Theorem 4.9 b).

**Lemma D.7** *Let $x \in [0, 1]$ and $\overline{v}, \overline{x} \in [0, \infty)$. For all probabilistic interpretations $Pr$ with $Pr(G) > 0$, the conditions $Pr(EG) = 0$, $\overline{v} \cdot Pr(G) = Pr(E\overline{G})$, $x \cdot Pr(G) = Pr(GF)$, and $\overline{x} \cdot Pr(G) = Pr(\overline{G}F)$ are equivalent to:*

| | | | | | | |
|---|---|---|---|---|---|---|
| $\frac{Pr(\overline{E}\,\overline{G}\,\overline{F})}{Pr(G)}$ | $\frac{Pr(\overline{E}\,\overline{G}F)}{Pr(G)}$ | $\frac{Pr(\overline{E}\,\overline{G})}{Pr(G)}$ | | $\frac{Pr(\overline{E}G\overline{F})}{Pr(G)}$ | $\frac{Pr(\overline{E}\,GF)}{Pr(G)}$ | 1 |
| $\frac{Pr(E\overline{G}\,\overline{F})}{Pr(G)}$ | $\frac{Pr(E\overline{G}F)}{Pr(G)}$ | $\overline{v}$ | | $\frac{Pr(EG\overline{F})}{Pr(G)}$ | $\frac{Pr(EGF)}{Pr(G)}$ | 0 |
| $\frac{Pr(\overline{G}\,F)}{Pr(G)}$ | $\overline{x}$ | | | $1 - x$ | $x$ | |

**Proof.** The claim can be verified by straightforward arithmetic transformations based on the properties of probabilistic interpretations. □

We are now ready to prove Theorem 4.9.

**Proof of Theorem 4.9.** a) By the definition of queries to conditional constraint trees, all paths from a basic event in $E$ to a basic event in $F$ have at least one basic event in common. Hence, we can choose the basic event $G$ from all such basic events in common such that $\exists (G|E)[z_1, z_2]$ is a strongly conclusion-restricted complete query to a subtree.

b) For $u_1 > 0$, the claim follows from Theorem 4.7 a) to c). For the special case of exact conditional constraint trees $(\mathcal{B}, KB)$, the claim then follows from Theorems 4.3 and 4.5.

Let $u_1 = 0$, $v_1 = 1$, and $G \Rightarrow F$. It holds $1 \cdot Pr(E) = Pr(EF)$ for all models $Pr$ of $KB$. Moreover, by Theorem 3.2, there exists a model $Pr$ of $KB$ with $Pr(E) > 0$.

Let $u_1 = 0$, $v_1 = 0$, and $G \Rightarrow F$. It is easy to see that by (1) and Theorem 4.7 d), the tight upper answer is given by $\{x_2/1\}$. We now show that the tight lower answer is given by $\{x_1/0\}$. By Theorem 4.3 b), there exists a model $Pr_1$ of $KB_1$ with $Pr_1(E) > 0$, $Pr_1(G) > 0$, and $Pr_1(EG) = 0$. By Theorem 3.2, there exists a model $Pr_2$ of $KB_2$ with $Pr_2(G) > 0$. By Lemma 3.1, we can choose $Pr_1$ and $Pr_2$ with $Pr_1(G) = Pr_2(G)$ and $Pr_1(\overline{E}\,\overline{G}) \geq Pr_2(\overline{G}F)$. By Lemmata A.1 and D.7, we can choose the probabilistic interpretation $Pr_0$ over $\{E, G, F\}$ with $Pr_0(A_1) = Pr_1(A_1)$ and $Pr_0(A_2) = Pr_2(A_2)$ for all atomic events $A_1$ and $A_2$ over $\{E, G\}$ and $\{G, F\}$, respectively, such that:

$$Pr_0(E\overline{G}F) = \max(0, Pr_2(\overline{G}F) - Pr_1(\overline{E}\,\overline{G})) = 0$$
$$Pr_0(EGF) = 0 .$$

By Lemma A.2, there exists a probabilistic interpretation $Pr$ over $\mathcal{B}$ with (12) for all atomic events $H_0$, $H_1$, $H_2$, $A_1$, and $A_2$ over the sets of basic events $\{G\}$, $\{E\}$, $\{F\}$, $\mathcal{B}_1 \setminus \{G\}$, and $\mathcal{B}_2 \setminus \{G\}$, respectively. Hence, $Pr$ is a model of $KB$ with $Pr(E) > 0$ and $Pr(EF) = 0$.

For $u_1 = 0$ and not $G \Rightarrow F$, the claim follows from (1) and Theorem 4.7 d). □